%% file: tr-seq.tex
\documentclass[12pt]{article}

\usepackage{latexsym}
\usepackage{amsmath}
\usepackage[usenames]{color}
\usepackage[dvips]{epsfig}
\usepackage{graphicx}

\input{newcommands}

\begin{document}

%\fontsize{11}{14.5pt}\selectfont

\begin{center}

%{\small Technical Report No.\ 07??, Department of Statistics, University of Toronto}

%\vspace*{0.8in}

{\large \bf A Method for Compressing Parameters in Bayesian Models with Application to Logistic Sequence Prediction Models\footnotemark[1]}\\[20pt]

{\large Longhai Li\footnotemark[2]} and {\large Radford M. Neal\footnotemark[3]}

%?? September 2007
 
\end{center}

\footnotetext[1]{This paper appeared as part of Longhai Li's PhD thesis (Li,
2007).}

\footnotetext[2]{Department of Mathematics and Statistics, University of
Saskatchewan, Saskatoon, Saskatchewan, S7N5E6, CANADA. Email:\texttt{longhai@math.usask.ca}}

\footnotetext[3]{Department of Statistics, University of
Toronto, Toronto, Ontario, M5S3G3, CANADA. Email: \texttt{radford@utstat.toronto.edu}}

\vspace{10pt}

\noindent {\bf Abstract.} Bayesian classification and regression with high order
interactions is largely infeasible because Markov chain Monte Carlo (MCMC) would
need to be applied with a great many parameters, whose number increases
rapidly with the order. In this paper we show how to make it feasible by
effectively reducing the number of parameters, exploiting the fact that many
interactions have the same values for all  training cases. Our method uses a
single ``compressed'' parameter to represent the sum of all parameters
associated with a set of patterns that have the same value for all training
cases. Using symmetric stable distributions as the priors of  the original
parameters, we can easily find the priors of these compressed parameters. We
therefore need to deal only  with a much smaller number of compressed parameters
when training the model with MCMC. The number of compressed parameters may have
converged before considering the highest possible order. After training the
model, we can split these compressed parameters into the original ones as needed
to make predictions for test cases. We show in detail how to compress parameters
for logistic sequence prediction models. Experiments on both simulated and real
data demonstrate that a huge number of parameters can  indeed be reduced by our
compression method.

\section{Introduction}

In many classification and regression problems, the response variable $y$
depends on  high-order interactions of ``features'' (also called ``covariates'',
``inputs'', ``predictor variables'', or ``explanatory variables''). Some complex
human diseases are found to be related to high-order interactions of
susceptibility genes and environmental exposures (Ritchie et. al. 2001).  The
prediction of the next character in English text  is improved by using a large
number of preceding characters (Bell, Cleary and Witten 1990). Many biological
sequences have long-memory properties.

When the features are discrete, we can employ high-order interactions in
classification and regression models by introducing, as additional predictor
variables, the indicators for each possible interaction pattern, equal to $1$ if
the pattern  occurs for a subject and $0$ otherwise. In this paper we will use
``features'' for the original discrete measurements and ``predictor variables''
for these derived variables, to distinguish them. The number of such predictor
variables increases exponentially with the order of interactions. The total
number of order-$k$ interaction patterns with $k$ binary (0/1) features is
$2^k$, accordingly we will have $2^k$ predictor variables. A model with
interactions of even a moderate order is prohibitive in real applications, 
primarily for computational reasons. People are often forced to use a model with
very small order, say only $1$ or $2$, which, however, may omit useful
high-order predictor variables.

Besides the computational considerations, classification and regression with a
great many predictor variables may ``overfit'' the data. Unless the number of
training cases is much larger than the number of predictor variables the model
may fit the noise instead of the signal in the data, with the result that
predictions for new test cases are poor. This problem can be solved by using
Bayesian modeling with appropriate prior distributions. In a Bayesian model, we
use a probability distribution over parameters to express our prior belief about
which configurations of parameters may be appropriate. One such prior belief is
that a parsimonious model can approximate the reality well. In particular, we
may believe that most high-order interactions are largely irrelevant to
predicting the response. We express such a prior by assigning each regression
coefficient a distribution with mode $0$, such as a Gaussian or Cauchy
distribution centered at $0$. Due to its heavy tail, a Cauchy distribution may
be more appropriate than a Gaussian distribution to express the prior belief
that almost all coefficients of high order interactions are close to $0$, with a
very small number of exceptions. Additionally, the priors we use for the widths
of Gaussian or Cauchy distributions for higher order interaction should favor
small values. The resulting joint prior for all coefficients favors a model with
most coefficients close to $0$, that is, a model emphasizing  low order
interactions. By incorporating such prior information into our inference, we
will not overfit the data with an unnecessarily complex model. 

However, the computational difficulty with a huge number of parameters is even
more pronounced for a Bayesian approach than other approaches, if we have to use
Markov chain Monte Carlo methods to sample from the posterior distribution,
which is computationally burdensome even for a moderate number of parameters.
With more parameters, a Markov chain sampler will take longer for each iteration
and require more memory, and may need more iterations to converge or get trapped
more easily in local modes. Applying Markov chain Monte Carlo methods to
classification and regression with high-order interactions therefore seems
infeasible.

In this paper, we show how these problems can be solved by effectively
reducing the number of parameters in a Bayesian model with high-order
interactions, using the fact that in a model that uses all interaction patterns,
from a low order to a high order, many predictor variables have the same values
for all training cases. For example, if an interaction pattern occurs in only
one training case, all the interaction patterns of higher order contained in it
will also occur in only that case and have the same values for all training
cases --- $1$ for that training case and $0$ for all others. Consequently, only
the sum of the coefficients associated with these predictor variables matters in
the likelihood function. We can therefore use a single ``compressed'' parameter
to represent the sum of the regression coefficients for a group of predictor
variables that have the same values in training cases. For models with very high
order of interactions, the number of such compressed parameters will be much
smaller than the number of original parameters. If the priors for the original
parameters are symmetric stable distributions, such as Gaussian or Cauchy, we
can easily find the prior distributions of these compressed parameters, as they
are also  symmetric stable distributions of the same type. In training the model
with Markov chain Monte Carlo methods we need to deal only with these compressed
parameters.  After training the model, the compressed parameters can be  split
into the original ones as needed to  make predictions for test cases.  Using our
method for compressing parameters, one can handle Bayesian regression and
classification problems with very high order of interactions in a reasonable
amount of time.

This paper will be organized as follows. In Section~\ref{sec-all-comp} we
describe in general terms the method of compressing parameters, and how to split
them to make predictions for test cases. We then apply the method to logistic
sequence models in Section~\ref{sec-blsm}. There, we will describe the specific
schemes for compressing parameters for the sequence prediction models, and use
simulated data and real data to demonstrate our method. We draw conclusions and
discuss future work in Section~\ref{sec-comp-conclude}. 

The software package (using R as interface but with most functions written in C) for the method described in this paper is available from \texttt{http://math.usask.ca/$\sim$longhai}.

\section{Our Method for Compressing Parameters}\label{sec-all-comp}

\subsection{Compressing Parameters} \label{sec-comp}

Our method for compressing parameters is applicable when we can divide the
regression coefficients used in the likelihood function into a number of groups
such that the likelihood is a function only of the sums over these groups. The
groups will depend on the particular training data set. An example of such a
group is the regression coefficients for a set of predictor variables that have
the same values for all training cases. It may not be easy to find an efficient
scheme for grouping the parameters of a specific model. We will describe how to
group the parameters for sequence prediction models in Section~\ref{sec-blsm}.
Suppose the number of such groups is $G$. The parameters in group $g$ are
denoted by $\beta_{g1},\ldots,\beta_{g,n_g}$, and the sum of them is denoted by
$s_g$:

\beq s_g=\sum_{k=1}^{n_g}\beta_{gk},\ \ \ \ \ \mbox{for }g=1,\ldots,G
\label{eqn-sg}
\eeq

\noindent We assume that the likelihood function can be written as:

\beq  \lefteqn{L^\beta(\beta_{11},\ldots,\beta_{1,n_1},\lldots,
               \beta_{G1},\ldots,\beta_{G,n_G})}\nonumber\\
&=&
L\left(\sum_{k=1}^{n_1}\beta_{1k},\,\ldots,\, \sum_{k=1}^{n_G}\beta_{Gk}
\right)
= L(s_1,\lldots,s_G) \label{eqn-like-train}
\eeq

\noindent  Note that the above $\beta$'s are only the regression coefficients
for the interaction patterns occurring in training cases. The predictive
distribution for a test case may use extra regression coefficients, whose
distributions depend only on the priors given relevant hyperparameters.

We need to define priors for the $\beta_{gk}$ in a way that lets us  easily find
the priors of the $s_g$. For this purpose, we assign each $\beta_{gk}$ a
symmetric stable distribution centered at $0$ with  width parameter
$\sigma_{gk}$. Symmetric stable distributions (Feller 1966) have the following
additive property: If random variables $X_1,\ldots,X_n$ are independent and have
symmetric stable distributions of index $\alpha$,  with location parameters $0$
and width parameters $\sigma_1,\ldots,\sigma_n$, then the sum of these random
variables, $\sum_{i=1}^n X_i$, also has a symmetric stable distribution of index
$\alpha$, with location parameter $0$ and width parameter $(\sum_{i=1}^n
\sigma_i^\alpha)^{1/\alpha}$. Symmetric stable distributions exist and are
unique for $\alpha\in (0,2]$. The symmetric stable distributions with
$\alpha=1$  are Cauchy distributions. The density function of a Cauchy
distribution with location parameter $0$ and width parameter $\sigma$ is
$[\pi\sigma(1+x^2/\sigma^2)]^{-1}$. The symmetric stable distributions with
$\alpha=2$ are Gaussian distributions, for which the width parameter is the
standard deviation. Since the symmetric stable distributions with $\alpha$ other
than $1$ or $2$ do not have closed form density functions,  we will use only 
Gaussian or Cauchy priors. That is, each parameter $\beta_{gk}$ has a Gaussian
or Cauchy distribution with location parameter $0$ and width parameter
$\sigma_{gk}$:

\beq \beta_{gk} \sim N(0,\sigma_{gk}^2)\ \ \ \mbox{or} \ \ \
\beta_{gk} \sim \cc(0,\sigma_{gk})  \label{eqn-comp-prior}
\eeq

\noindent Some $\sigma_{gk}$ may be common for different $\beta_{gk}$, but for
the moment we denote them individually. We might also treat the $\sigma_{gk}$'s
as unknown hyperparameters, but again we assume them fixed for the moment.

If the prior distributions for the $\beta_{gk}$'s are as
in~(\ref{eqn-comp-prior}), the prior distribution of $s_g$ can be found using
the property of symmetric stable distributions:

\beq
s_g \sim N\left(0,\ \sum_{k=1}^{n_g}\sigma_{gk}^2\right)\ \ \ \mbox{or}\ \ \
s_g \sim \cc\left(0,\ \sum_{k=1}^{n_g}\sigma_{gk}\right)
\label{eqn-dist-sg}
\eeq

Let us denote the density of $s_g$ in~(\ref{eqn-dist-sg}) by $P_g^s$ (either a
Gaussian or Cauchy), and denote $s_1,\ldots,s_G$ collectively by $\mb s$. The
posterior distribution can be written as follows:

\beq P(\mb s \given \trainingdata)= {1\over
c(\trainingdata)}\,L(s_1,\lldots,s_G)\ P_1\sp{s}(s_1)\ \cdots \ P_g\sp{s}(s_G)
\label{eqn-post-sum} \eeq

\noindent where $\trainingdata$ is the training data, and $c(\trainingdata)$ is
the marginal probability or density function of $\trainingdata$.

Since the likelihood function $L(s_1,\lldots,s_G)$ typically depends on
$s_1,\ldots,s_G$ in a complicated way, we may have to use some Markov chain
sampling method to sample for $\mb s$ from distribution~(\ref{eqn-post-sum}). 

\subsection{Splitting Compressed Parameters}\label{sec-split}

\begin{figure}[t]

\begin{center} \includegraphics[width=5.0in,height=1in]{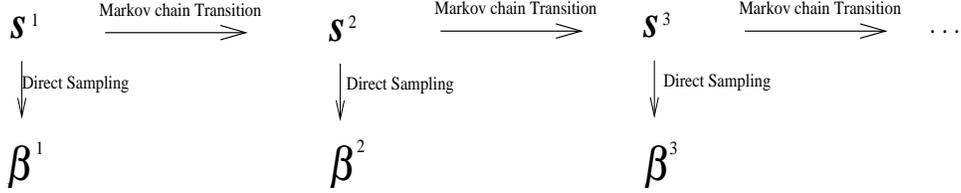}
\end{center}

\caption{A picture depicting the sampling procedure after compressing parameters.}

\label{fig-sample-comp}

\end{figure}

After we have obtained samples of $s_g$, probably using some Markov chain
sampling method, we may need to split them into their original components
$\beta_{g1},\ldots,\beta_{g,n_g}$ to make predictions for test cases. This
``splitting'' distribution depends only on the prior distributions, and is
independent of the training data $\trainingdata$. In other words, the splitting
distribution is just the conditional distribution of
$\beta_{g1},\ldots,\beta_{gn_g}$ given $\sum_{k=1}^{n_g}\beta_{gk}=s_g$, whose
density function is:

\beq
P(\beta_{g1},\ldots,\beta_{g,n_g-1}\given s_g) =
\left[\prod_{k=1}^{n_g-1}\ P_{gk}(\beta_{gk})\right]\
P_{g,n_g}\left(s_g - \sum_{k=1}^{n_g-1}\beta_{gk}\right)\,/\,P_g^s(s_g)
\label{eqn-split1}
\eeq

\noindent where $P_{gk}$ is the density function of the prior for $\beta_{gk}$.
Note that $\beta_{g,n_g}$ is omitted since it is equal to $s_g -
\sum_{k=1}^{n_g-1}\beta_{gk}$. 

As will be discussed in the Section~\ref{sec-split-pred}, sampling
from~(\ref{eqn-split1}) can be done efficiently by a direct sampling method,
which does not involve costly evaluations of the likelihood function. We need to
use Markov chain sampling methods and evaluate the likelihood function only when
sampling for $\mb s$. Figure~\ref{fig-sample-comp} shows the sampling procedure
after compressing parameters, where $\mb \beta$ is a collective representation
of $\beta_{gk}$, for $g=1,\ldots,G, k = 1,\ldots,n_g-1$. When we consider
high-order interactions, the number of groups, $G$, will be much smaller than
the number of $\beta_{gk}$'s. This procedure is therefore much more efficient
than  applying Markov chain sampling methods to all the original $\beta_{gk}$
parameters.

Furthermore, when making predictions for a particular test case, we actually do
not need to sample from the distribution~(\ref{eqn-split1}), of dimension
$n_g-1$, but only from a derived 1-dimensional distribution, which saves a huge
amount of space. 

Before discussing how to sample from~(\ref{eqn-split1}), we first phrase this
compressing-splitting procedure more formally in the next section to show its
correctness. 

\subsection{Correctness of the Compressing-Splitting Procedure}
\label{sec-correctness}

The above procedure of compressing and splitting parameters can be seen in terms
of a transformation of the original parameters $\beta_{gk}$  to a new set of
parameters containing $s_g$'s, as defined in~(\ref{eqn-sg}), in light of the
training data. The posterior distribution~(\ref{eqn-post-sum}) of $\mb s$ and
the splitting distribution~(\ref{eqn-split1}) can be derived from the joint
posterior distribution of the new parameters. 

The invertible mappings from the original parameters to the new parameters are
shown as follows, for $g=1,\ldots,G$,

\beq 
(\beta_{g1},\ldots,\beta_{g,n_g-1},\beta_{g,n_g})\ \ \Longrightarrow \ \ 
(\beta_{g1},\ldots,\beta_{g,n_g-1},\sum_{k=1}^{n_g}\beta_{gk}) =
 (\beta_{g1},\ldots,\beta_{g,n_g-1},s_g)
\label{eqn-transform}
\eeq

\noindent In words, the first $n_g-1$ original parameters $\beta_{gk}$'s are
mapped to themselves (we might use another set of symbols, for example $b_{gk}$,
to denote the new parameters, but here we still use the old ones for simplicity
of presentation while making no confusion), and the sum of all $\beta_{g,k}$'s,
is mapped to $s_g$.  Let us denote the new parameters $\beta_{gk}$, for
$g=1,\ldots,G, k = 1,\ldots,n_g-1$, collectively by $\mb \beta$, and denote
$s_1,\ldots,s_g$ by $\mb s$. (Note that $\mb \beta$ does not include
$\beta_{g,n_g}$, for $g=1,\ldots,G$. Once we have obtained the samples of $\mb
s$ and $\mb \beta$ we can use $\beta_{g,n_g}=s_g - \sum_{k=1}^{n_g-1}\beta_{gk}$
to obtain the samples of $\beta_{g,n_g}$.)

The posterior distribution of the original parameters, $\beta_{gk}$, is:

\beq
P(\beta_{11},\lldots,\beta_{G,n_G}\given \trainingdata)
=
{1\over c(\trainingdata)}
L\left(\sum_{k=1}^{n_1}\beta_{1k},\,\ldots,\,
  \sum_{k=1}^{n_G}\beta_{Gk}\right)
\prod_{g=1}^G\prod_{k=1}^{n_g}\ P_{gk}(\beta_{gk})
\label{eqn-original-post}
\eeq

\noindent By applying the standard formula for the density function of
transformed random variables, we can obtain from~(\ref{eqn-original-post}) the
posterior distribution of the $\mb s$ and $\mb \beta$:

\beq
P(\mb s,\mb \beta \given \trainingdata ) =
{1\over c(\trainingdata)}\,
L\left(s_1,\,\ldots,\,s_G\right)
\prod_{g=1}^G\left[
\prod_{k=1}^{n_g-1}P_{gk}(\beta_{gk})\right]
P_{g,n_g}\left(s_g - \sum_{k=1}^{n_g-1}\beta_{gk}\right)
\,|\det(J)|
\label{eqn-transform-post}
\eeq

\noindent where the $|\det(J)|$ is absolute value of the determinant of the
Jacobian matrix, $J$, of the mapping~(\ref{eqn-transform}), which can be shown
to be $1$.

Using the additive property of symmetric stable distributions, which is stated
in section~\ref{sec-comp}, we can analytically integrate out $\mb \beta$ in
$P(\mb s,\mb \beta \given \trainingdata)$, resulting in the marginal
distribution $P(\mb s \given \trainingdata)$:

\beq
P(\mb s \given \trainingdata)
&=& \int\,P(\mb s,\mb \beta \given \trainingdata )\,d\mb\beta \\
&=& {1\over c(\trainingdata)}\,
L\left(s_1,\,\ldots,\,s_G\right)\nonumber \cdot\\
&&\ \ \ \prod_{g=1}^G\int\cdots\int\,\left[
\prod_{k=1}^{n_g-1}P_{gk}(\beta_{gk})\right]
P_{g,n_g}\left(s_g - \sum_{k=1}^{n_g-1}\beta_{gk}
\right)d\beta_{g1}\cdots d\beta_{g,n_g-1}\\
&=& {1\over c(\trainingdata)}\,L\left(s_1,\,\ldots,\,s_G\right)\,
    P^s_1(s_1)\ \cdots\ P^s_G(s_G)
\label{eqn-dist-s}
\eeq

The conditional distribution of $\mb\beta$ given $\trainingdata$ and $\mb s$ can
then be obtained as follows:

\beq
P(\mb \beta \given \mb s,\trainingdata)
& = & P(\mb s,\mb \beta \given \trainingdata)\,/\,P(\mb
s \given \trainingdata) \\
& = & \prod_{g=1}^G\,\left[
\prod_{k=1}^{n_g-1}P_{gk}(\beta_{gk})\right]
P_{g,n_g}\left(s_g - \sum_{k=1}^{n_g-1}\beta_{gk}\right)\,/\,P^s_g(s_g)
\label{eqn-cond-beta-s}
\eeq

\noindent From the above expression, it is clear that $P(\mb \beta \given \mb
s,\trainingdata)$ is unrelated to $\trainingdata$, i.e., $P(\mb \beta\given \mb
s,\trainingdata)=P(\mb \beta \given \mb s)$, and is independent for different
groups. Equation~(\ref{eqn-split1}) gives this distribution only for one group
$g$.

\subsection{Sampling from the Splitting Distribution}
\label{sec-split-pred}

In this section, we discuss how to sample from the splitting
distribution~(\ref{eqn-split1}) to make predictions for test cases after we have
obtained samples of $s_1,\ldots,s_G$. 

If we sampled for all the $\beta_{gk}$'s, storing them would require a huge
amount of space when the number of parameters in each group is huge. We
therefore sample for $\mb \beta$ conditional on $s_1,\ldots,s_G$ only
temporarily, for a particular test case. As will be seen in
Section~\ref{sec-blsm}, the predictive function needed to
make prediction for a particular test case, for example the probability that a
test case is in a certain class, depends only on the sums of subsets of
$\beta_{gk}$'s in  groups. After re-indexing the $\beta_{gk}$'s in each group
such that the $\beta_{g1},\ldots,\beta_{g,t_g}$ are those needed by the test
case, the variables needed for making a prediction for the test case are:

\beq  
s^t_g &=& \sum_{k=1}^{t_g}\beta_{gk}\, , \mbox{ for } g=1,\ldots,G,
\eeq

\noindent Note that when $t_g=0$, $s^t_g=0$, and when $t_g=n_g$, $s^t_g = s_g$.
The predictive function may also use some sums of  extra regression coefficients
associated with the interaction patterns that occur in this test case but not in
training cases. Suppose  the extra regression coefficients need to be divided
into $Z$ groups, as required by the form of the predictive function, which we
denote by  $\beta_{11}^*,\ldots,\beta_{1,n^*_1}^*,\ldots,\beta_{Z,1}^*,
\ldots,\beta_{Z,n^*_Z}^*$. The variables needed for making prediction for the
test cases are:

\beq
s^*_z
&=& \sum_{k=1}^{n^*_z}\beta_{zk}^*\, , \mbox{ for } z =1,\ldots,Z \eeq

In terms of the above variables, the function needed to make a prediction for a
test case can be written as

\beq a\left(\sum_{k=1}^{t_1}\beta_{1k},\ \ldots \ ,\sum_{k=1}^{t_G}\beta_{Gk},\
\sum_{k=1}^{n^*_1}\beta_{1k}^*,\lldots,\sum_{k=1}^{n^*_Z}\beta_{Zk}^* \right)
\label{eqn-predfun} = a(s_1^t,\lldots,s_G^t,s_1^*,\lldots,s_Z^*) 
\label{eqn-blsm-pred} \eeq

\noindent 

Let us write $s^t_1,\ldots,s^t_G$ collectively as $\mb s^t$, and write
$s^*_1,\ldots,s^*_Z$ as $\mb s^*$. The integral required to make a prediction
for this test case is

\beq \int\ a(\mb s^t,\mb s^*)\ P(\mb s^*)\ P(\mb s \given \trainingdata)\
\prod_{g=1}^G\ P(s^t_g\given s_g)\ d\mb s\ d\mb s^t d\mb s^*.
\label{eqn-int-pred} \eeq

The integral over $\mb s^t$ is done by MCMC. We also need to sample for $\mb
s^*$ from $P(\mb s^*)$, which is the prior distribution of $\mb s^*$ given some
hyperparameters (from the current MCMC iteration) and can therefore be sampled
easily. Finally, we need to sample from $P(s^t_g\given s_g)$, which can be
derived from~(\ref{eqn-split1}), shown as follows: 

\beq P(s^t_g \given s_g) = P^{(1)}_g(s_g^t)\
P^{(2)}_g(s_g-s_g^t)\,/\,P^s_g(s_g) \label{eqn-split2} \eeq

\noindent where $P^{(1)}_g$ and  $P^{(2)}_g$ are the priors (either Gaussian or
Cauchy) of $\sum_{1}^{t_g}\beta_{gk}$ and $\sum_{t_g+1}^{n_g}\beta_{gk}$,
respectively. We can obtain~(\ref{eqn-split2}) from~(\ref{eqn-split1})
analogously as we obtained the density of $s_g$, that is, by first mapping $\mb
\beta$ and $\mb s$ to a set of new parameters containing $\mb s$ and $\mb s^t$,
then integrating away other parameters, using the additive property of
symmetric stable distributions. The distribution~(\ref{eqn-split2}) splits 
$s_g$ into two components.

When the priors for the $\beta_{gk}$'s are Gaussian distributions, the
distribution~(\ref{eqn-split2}) is also a Gaussian distribution, given as
follows:

\beq s^t_g \given s_g \ \sim \  N\left(s_g\
\frac{\sigmaa^2}{\sigmaa^2 +
\sigmab^2}\ ,\  \sigmaa^2\left(1\ -\
\frac{\sigmaa^2}{\sigmaa^2 +
\sigmab^2}\right) \right)
\label{eqn-cond-gs}
\eeq

\noindent where $\sigmaa^2 = \sum_{k=1}^{t_g}\sigma_{gk}^2$
and $\sigmab^2 = \sum_{t_g+1}^{n_g}\sigma_{gk}^2$.
Since~(\ref{eqn-cond-gs}) is a Gaussian distribution,  we can  sample from it
by standard methods. 

When we use Cauchy distributions as the priors for the $\beta_{gk}$'s, the
density function of~(\ref{eqn-split2}) is:

\beq
P(s_g^t \given s_g) = {1\over C}\,
 \frac{1}{ \sigmaa^2 + (s_g^t)^2}\
 \frac{1}{ \sigmab^2 + (s_g^t-s_g)^2}
\label{eqn-cond-cauchy}
\eeq

\noindent where $\sigmaa = \sum_{k=1}^{t_g}\sigma_{gk}$, $\sigmab  =
\sum_{t_g+1}^{n_g}\sigma_{gk}$, and $C$ is the normalizing constant given below
by~(\ref{eqn-C-cauchy}).

When $s_g=0$ and $\sigmaa=\sigmab$, the distribution~(\ref{eqn-cond-cauchy}) is
a t-distribution with $3$ degrees of freedom, mean $0$ and width
$\sigmaa/\sqrt{3}$, from which we can sample by standard methods. Otherwise, the
cumulative distribution function (CDF) of~(\ref{eqn-cond-cauchy}) can be shown
to be:

\beq
F(\st\,;\,\s,\sigmaa,\sigmab)
& = &\frac{1}{C}\,
     \left[r
           \log\left(\frac{(\st)^2+\sigmaa^2}
                          {(\st-\s)^2+\sigmab^2}
               \right) + \right.
     \nonumber\\
&   &\ \ \ \ \ \ \
      p_0\,\left(\arctan\left(\frac{\st}{\sigmaa}\right)+\frac{\pi}{2}\right)+
     \nonumber\\
&   &\ \ \ \ \ \ \
     \left.
           p_s\,\left(\arctan\left(\frac{\st-\s}{\sigmab}\right)+\frac{\pi}{2}
                \right)
     \right]
\label{eqn-cdf-split-cc1}
\eeq

\noindent where

\beq
C  &=&{\pi\,(\sigmaa+\sigmab)
       \over \sigmaa\sigmab\,(\s^2+(\sigmaa+\sigmab)^2)}\,,
   \label{eqn-C-cauchy}\\
r  &=&\frac{\s}{\s^4+2\left(\sigmaa^2+\sigmab^2\right)\,\s^2+
      \left(\sigmaa^2-\sigmab^2\right)^2}\,, \\
p_0&=&{1\over \sigmaa}\,\frac{\s^2 -
      \left(\sigmaa^2-\sigmab^2\right)}
      {\s^4+2\left(\sigmaa^2+\sigmab^2\right)\,\s^2+
      \left(\sigmaa^2-\sigmab^2\right)^2},\\
p_s&=&{1\over \sigmab}\,\frac{\s^2 +
      \left(\sigmaa^2-\sigmab^2\right)}
      {\s^4+2\left(\sigmaa^2+\sigmab^2\right)\,\s^2+
      \left(\sigmaa^2-\sigmab^2\right)^2}
\eeq

When $s_g\not=0$, the derivation of~(\ref{eqn-cdf-split-cc1})  uses the
equations below from~(\ref{eqn-decomp1-b}) to~(\ref{eqn-decomp1-e}) as follows,
where $p=(a^2-c)/b,q=b+q,r=pc-a^2q$, and we assume $4c-b^2>0$, 

\beq
\frac{1}{x^2+a^2}\,\frac{1}{x^2+bx+c}
&\hspace*{-0.1in}=&\hspace*{-0.1in}
{1\over r}\,\left(\frac{x+p}{x^2+a^2} - \frac{x+q}{x^2+bx+c}
        \right) \label{eqn-decomp1-b}\\
\int_{-\infty}^x\frac{u+p}{u^2+a^2}du
&\hspace*{-0.1in}=&\hspace*{-0.1in}
{1\over2}\,\log(x^2 + a^2) + {p\over a}\,\arctan\left({x\over a}\right)
     +{\pi\over 2}
\\
\int_{-\infty}^x{u+q \over u^2+bu+c}du
&\hspace*{-0.1in}=&\hspace*{-0.1in}
{1\over2}\,\log(x^2 + bx + c) +
{2q-b \over \sqrt{4c-b^2}}\,\arctan\left({2x+b \over \sqrt{4c-b^2}}\right)
+{\pi\over 2} \label{eqn-decomp1-e}
\eeq

When $s_g=0$, the derivation of~(\ref{eqn-cdf-split-cc1}) uses the following
equations:

\beq
\frac{1}{x^2+a^2}\,\frac{1}{x^2+b^2} &=& 
\frac{1}{b^2-a^2}\,\left(\frac{1}{x^2+a^2}-\frac{1}{x^2+b^2}\right)\\
\int_{-\infty}^x \frac{1}{u^2+a^2}\,du &=& 
\frac{1}{a}\left(\arctan\left({x\over a}\right)+\frac{\pi}{2}\right)
\eeq

Since we can compute the CDF of~(\ref{eqn-cond-cauchy}) with 
~(\ref{eqn-cdf-split-cc1}) explicitly, we can use the inversion method to sample
from~(\ref{eqn-cond-cauchy}), with the inverse CDF computed by some numerical
method. We chose the Illinois method (Thisted 1988, Page 171), which is robust
and fairly fast.

When sampling for $s^t_1,\ldots,s^t_G$ temporarily for each test case is not
desired, for example, when we need to make predictions for a huge number of test
cases at a time, we can still apply  the above method that splits a Gaussian or
Cauchy random variable into two parts $n_g-1$ times to split $s_g$  into $n_g$
parts. Our method for compressing parameters is still useful because sampling
from the splitting distributions uses direct sampling methods, which are much
more efficient than applying Markov chain sampling method to the original
parameters. However, we will not save space if we take this approach of sampling
for all $\beta$'s.

\section{Application to Sequence Prediction Models} \label{sec-blsm}

In this section, we show how to compress parameters of logistic sequence
prediction models in which states of a sequence  are discrete. We will first
define this class of models, and then describe the scheme for grouping the
parameters. To demonstrate our method, we use a binary data set generated using
a hidden Markov model, and a data set created from English text, in which each
state has 3 possibilities (consonant, vowel, and others). These experiments show
that our compression method produces a large reduction in the number of
parameters needed for training the model, when the prediction for the next state
of a sequence is based on a long  preceding sequence, i.e., a high-order model.
We also show that good predictions on test cases result from being able to use a
high-order model.

\subsection{Bayesian Logistic Sequence Prediction Models}
\label{sec-def-blsm}

\begin{figure}[t]

\begin{center}

\includegraphics[scale=0.85]{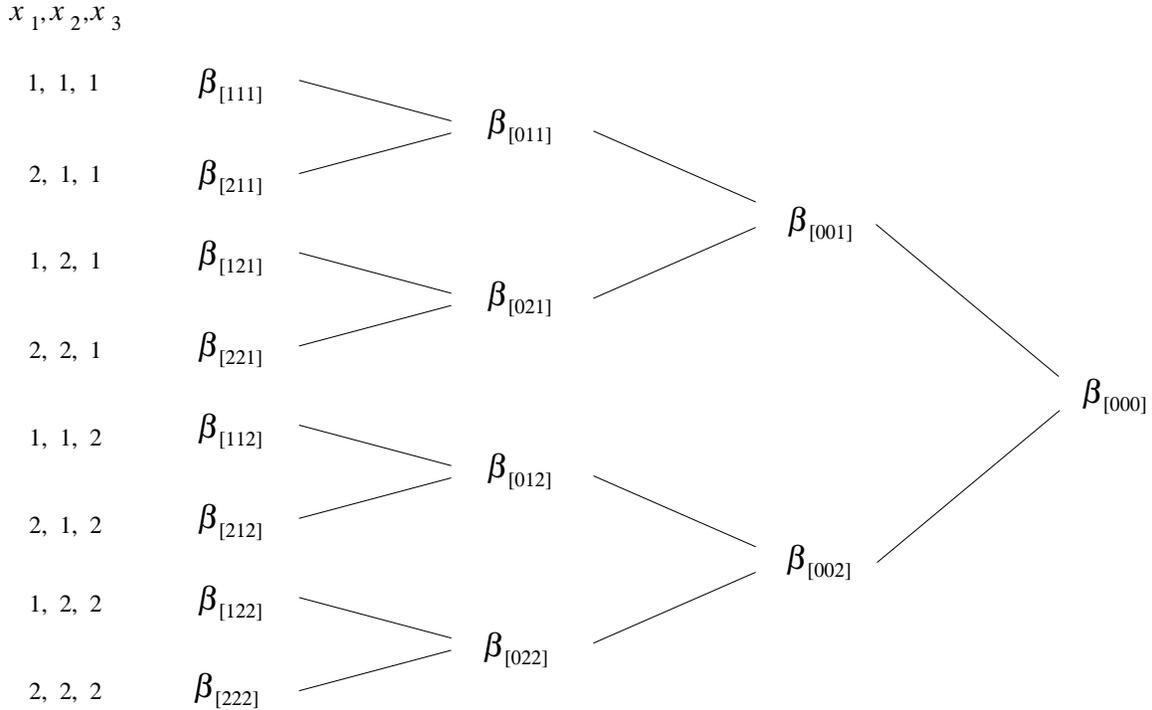}

\end{center}

\caption[A picture of the coefficients, $\mb \beta$, for all patterns  in 
binary sequences of length $O=3$.]{ A picture of the coefficients, $\mb \beta$,
for all patterns  in  binary sequences of length $O=3$. $\beta_{[A_1A_2A_3]}$ is
associated with the pattern written as $[A_1A_2A_3]$, with $A_t=0$ meaning that
$x_t$ is allowed to be either $1$ or $2$, in other words, $x_t$ is ignored in
defining this pattern. For example, $\beta_{[000]}$ is the intercept term. These
coefficients are used in defining the linear function $l\,((x_1,x_2,x_3),\mb
\beta)$ in the logistic model~(\ref{eqn-softmax}). For each combination of
$(x_1,x_2,x_3)$ on the left column, $l\,((x_1,x_2,x_3),\mb \beta)$ is equal to
the sum of $\beta$'s along the path linked by lines, from $\beta_{[x_1 x_2
x_3]}$ to $\beta_{[000]}$. }

\label{fig-seq}

\end{figure}

We write a sequence of length $O+1$ as $x_1,\ldots,x_O,x_{O+1}$, where $x_t$
takes values from $1$ to $K_t$, for $t=1,\ldots,O$, and $x_{O+1}$ takes values
from $1$ to $K$. We call $x_1,\ldots,x_O = \x_{1:O}$ the historic sequence. For
subject $i$ we write its historic sequence and response as $\mb x_{1:O}^{(i)}$
and $x^{(i)}_{O+1}$. We are interested in modelling the conditional distribution
$P(x_{O+1}\given \mb x_{1:O})$.

An interaction pattern $\P$ is written as $[A_1A_2\ldots A_O]$, where $A_t$ can
be from $0$ to $K_t$, with $A_t=0$ meaning that $x_t$ can be any value from $1$
to $K_t$. For example, $[0\ldots 01]$ denotes the pattern that fixes $x_O=1$ and
allows $x_1,\ldots,x_{O-1}$ to be any values in their ranges.  When all nonzero
elements of $\P$ are equal to the corresponding elements of a historic sequence,
$\mb x_{1:O}$, we say that pattern $\P$ occurs in $\mb x_{1:O}$, or pattern $\P$
is expressed by $\mb x_{1:O}$, denoted by $\mb x_{1:O}\in \P$. We will use  the
indicator $I(x_{1:O}\in \P)$ as a predictor variable, whose coefficient is
denoted by $\beta_{\P}$. For example, $\beta_{[0\cdots0]}$ is the intercept
term. A logistic model assigns each possible value of the response a linear
function of the predictor variables. We use $\beta^{(k)}_{\P}$ to denote the
coefficient associated with pattern $\P$ and used in the linear function for 
$x_{O+1} = k$.

For modeling sequences, we consider only the patterns where all zeros (if any)
are at the start.  Let us denote all such patterns by $\mb {\mathcal{S}}$. We
write all coefficients for $x_{O+1}=k$, i.e.,  $\left\{\beta^{(k)}_{\P} \given
\P \in \mb {\mathcal{S}}\right\}$, collectively as $\mb \beta^{(k)}$.
Figure~(\ref{fig-seq}) displays $\mb \beta^{(k)}$  for binary sequence of length
$O=3$, for some $k$, placed in a tree-shape.

Conditional on $\mb \beta^{(1)},\ldots,\mb \beta^{(K)}$ and $\mb x_{1:O}$, the
distribution of $x_{O+1}$ is defined as

\beq
P(x_{O+1} = k \given \mb x_{1:O}, \mb \beta^{(1)},\ldots,\mb \beta^{(K)} ) =
\frac{\exp( l\,(\x_{1:O},\mb\beta^{(k)}) ) }
     {\sum_{j=1}^K \exp( l\,(\x_{1:O},\mb\beta^{(j)}) ) }
\label{eqn-softmax}
\eeq

\noindent where

\beq
l\,(\x_{1:O},\mb\beta^{(k)})
&=& \sum_{\P \in \mb {\mathcal{S}}}\beta^{(k)}_{\P}\ I(\x_{1:O}\in \P)=
     \beta^{(k)}_{[0\cdots0]} +
    \sum_{t=1}^O \beta^{(k)}_{[0\cdots x_t\cdots x_O]}
\label{eqn-seq-linear}
\eeq

In Figure~\ref{fig-seq}, we display the linear functions for each possible
combination of $(x_1,x_2,x_3)$ on the left column, by linking together all
$\beta$'s in the summation~(\ref{eqn-seq-linear})  with lines, from
$\beta_{[x_1x_2x_3]}$ to $\beta_{[000]}$.

The prior for each $\beta_{\P}^{(k)}$ is a Gaussian or Cauchy distribution
centered at $0$, whose width depends on the order, $o(\P)$, of $\P$, which is
the number of nonzero elements of $\P$. There are $O+1$ such width parameters,
denoted by $\sigma_0,\ldots,\sigma_O$.  The $\sigma_o$'s are treated as
hyperparameters, assigned Inverse Gamma prior distributions with some shape
and rate parameters, leaving their values to be determined by the data. In
summary, the hierarchy of the priors is:

\beq \begin{array}{rcl} \sigma_o & \sim &
\mbox{Inverse-Gamma}(\alpha_o\,,(\alpha_o+1)\,w_o), \mbox{ for } o=0,\ldots,O\\
\beta^{(k)}_{\P}\given \sigma_{o(\P)} &\sim& \cc(0,\sigma_{o(\P)}) \mbox{ or }
N(0,\sigma_{o(\P)}^2), \mbox{ for } \P \in \mb {\mathcal{S}} \end{array}
\label{eqn-cc-prior} \eeq

\noindent where Inverse-Gamma$(\alpha,\lambda)$ denotes an Inverse Gamma
distribution with density function
$x^{-\alpha-1}\,\lambda^\alpha\,\exp(-\lambda/x)/\Gamma(\alpha)$. We express
$\alpha$ and $\lambda$  in~(\ref{eqn-cc-prior}) so that the mode of the prior is
$w_o$.

\subsection{Remarks on the Sequence Prediction Models} \label{sec-remark-blsm}

The Inverse Gamma distributions have heavy upward tails when $\alpha$ is small,
and particularly when $\alpha \leq 1$, they have infinite means. An Inverse
Gamma distribution with $\alpha_o\leq 1$ and small $w_o$, favors small values
around $w_o$, but still allows $\sigma_o$  to be exceptionally large, as needed
by the data. Similarly, the Cauchy distributions have heavy two-sided tails. The
absolute value of a Cauchy random variable has infinite mean. When a Cauchy
distribution with center $0$ and a small width is used as the prior for a group
of parameters, such as all $\beta$'s of the interaction patterns with the same
order in~(\ref{eqn-cc-prior}), a few parameters may be much larger in absolute
value than others in this group. As the priors for the coefficients of
high-order interaction patterns, the Cauchy distributions can therefore express
more accurately than the Gaussian distributions the prior belief that most
high-order interaction patterns are useless in predicting the response, but a
small number may be important.

It seems redundant to use a $\mb \beta^{(k)}$ for each $k=1,\ldots,K$
in~(\ref{eqn-softmax}) since only the differences between $\mb\beta^{(k)}$
matter in~(\ref{eqn-softmax}). A non-Bayesian model could fix one of them, say
$\mb \beta^{(1)}$, all equal to $0$, so as to make the parameters identifiable.
However, when $K\not=2$, forcing $\mb \beta^{(1)}=0$ in a Bayesian model will
result in a prior that is not symmetric for all $k$, which we may not be able to
justify. When $K=2$, we do require that $\mb \beta^{(1)}$ are all equal to $0$,
as there is no asymmetry problem.

Inclusion of $\beta_\P$ other than the highest order is also a redundancy, which
facilitates the expression of appropriate prior beliefs. The prior distributions
of linear functions of similar historic sequences $x_{1:O}$ are positively
correlated since they share some common $\beta$'s, for example, in the model
displayed by Figure~\ref{fig-seq}, $l\,((1,1,1),\mb \beta)$ and $l\,((2,1,1),\mb
\beta)$ share $\beta_{[011]},\beta_{[001]}$ and $\beta_{[000]}$. Consequently,
the predictive distributions of $x_O$ are similar given similar $x_{1:O}$.  By
incorporating such a prior belief into our inference, we borrow ``statistical
strength'' for those historic sequences with few replications in the training
cases from other similar sequences with more replications, avoiding making an
unreasonably extreme conclusion due to a small number of replications.

\subsection{Specifications of the Priors and Computation Methods}

\subsubsection{The Priors for the Hyperprameters}
\label{sec-priors-seq}

We fix $\sigma_0$ at $5$ for the Cauchy models and $10$ for the Gaussian models.
For $o>0$, the prior for $\sigma_o$ is Inverse
Gamma$(\alpha_o,(\alpha_o+1)w_o)$, where $\alpha_o$ and $w_o$ are:

\beq
\alpha_o=0.25,\ \ \ w_o = 0.1/o,\ \ \ \ \mbox{for }o =1,\ldots,O
\eeq

\noindent  The quantiles of Inverse-Gamma$(0.25,1.25\times 0.1)$, the prior of
$\sigma_1$, are shown as follows:

\noindent$$
\begin{array}{l|lllllllllll}
p & 0.01&0.1 & 0.2 & 0.3 &  0.4 &  0.5 &  0.6 & 0.7  & 0.8&0.9&0.99\\
\hline
q & 0.05&0.17& 0.34& 0.67&  1.33&  2.86&  7.13& 22.76& 115.65 &1851.83
  &1.85\times 10^7
\end{array}
$$

\noindent The quantiles of other $\sigma_o$ can be obtained by multiplying those
of $\sigma_1$ by $1/o$.

\subsubsection{The Markov Chain Sampling Method}

We use Gibbs sampling to sample for both the $s_g$'s (or the $\beta_{gk}$'s when
not applying our compression method) and the hyperparameters, $\sigma_o$. These 
1-dimensional conditional distributions are sampled using the slice sampling
method (Neal 2003), summarized as follows.  In order to sample from a
1-dimensional distribution with density $f(x)$, we can draw points $(x,y)$ from
the uniform distribution over the set $\{(x,y)\given 0 < y < f(x)\}$, i.e., the
region of the 2-dimensional plane between the x-axis and the curve of $f(x)$.
One can show that the marginal distribution of $x$ drawn this way is $f(x)$.  We
can use Gibbs sampling scheme to sample from the uniform distribution over
$\{(x,y)\given 0 < y < f(x)\}$. Given $x$, we can draw $y$ from the uniform
distribution over $\{y\given 0< y < f(x)\}$. Given $y$, we need to draw $x$ from
the uniform distribution over the ``slice'', $S=\{x\given f(x) > y\}$. However,
it is generally infeasible to draw a point directly from the uniform
distribution over $S$. Neal (2003) devises several Markov chain sampling schemes
that leave this uniform distribution over $S$ invariant. One can show that this
updating of $x$ along with the previous updating of $y$ leaves $f(x)$ invariant.
Particularly we chose the ``stepping out'' plus ``shrinkage'' procedures. The
``stepping out'' scheme first steps out from the point in the previous
iteration, say $x_0$, which is in $S$, by expanding an initial interval, $I$, of
size $w$ around $x_0$ on both sides with intervals of size $w$, until the ends
of $I$ are outside $S$, or the number of steps has reached a pre-specified
number, $m$. To guarantee correctness, the initial interval $I$ is positioned
randomly around $x_0$, and $m$ is randomly aportioned for the times of stepping
right and stepping left.  We then keep drawing a point uniformly from $I$ until
obtaining an $x$ in $S$. To facilitate the process of obtaining an $x$ in $S$,
we shrink the interval $I$ if we obtain an $x$ not in $S$ by cutting off the
left part or right part of $I$  depending on whether $x<x_0$ or $x>x_0$. 

We set $w=20$ when sampling for $\beta$'s if we use Cauchy priors, considering
that there might be two modes in this case, and set $w=10$ if we use Gaussian
priors. We set $w=1$ when sampling for $\sigma_o$. The value of $m$ is $50$ for
all cases. We trained the Bayesian logistic sequence model, with the compressed
or the original parameters, by running the Markov chain 2000 iterations, each
updating the $\beta$'s $1$ time, and updating the $\sigma$'s $10$ times, both
using slice sampling. The first $750$ iterations were discarded, and every $5$th
iteration afterward was used to predict for the test cases. 

The above specification of Markov chain sampling and the priors for the 
hyperparameters will be used for all experiments in this paper.

\subsection{Grouping Parameters of Sequence Prediction
Models}\label{compress-seq}

In this section, we describe a scheme for dividing the $\beta$'s  into a number
of groups, based on the training data, such that the likelihood function depends
only on the sums in groups, as shown by~(\ref{eqn-like-train}). The likelihood
function of $\mb\beta^{(k)}$, for $k=1,\ldots,K$, is the product of
probabilities in~(\ref{eqn-softmax}) applied to the training cases,
$\x^{(i)}_{1:O}, x^{(i)}_{O+1}$, for $i=1,\ldots,N$ (collectively denoted by
$\trainingdata$). It can be written as follows:

\beq
L^\beta(\mb\beta^{(1)},\ldots,\mb\beta^{(K)} \given \trainingdata) =
\prod_{i=1}^N
     \frac{\exp( l\,(\x^{(i)}_{1:O},\mb\beta^{(x^{(i)}_{O+1})}) ) }
     {\sum_{j=1}^K \exp( l\,(\x^{(i)}_{1:O},\mb\beta^{(j)}) ) }
\label{eqn-seq-like}
\eeq

\noindent Note that when $K=2$, $\mb\beta^{(1)}$ is fixed at $0$, and therefore
not included in the above likelihood function. But for simplicity, we do not
write another expression for $K=2$.

Since the linear functions with different $k$'s have the same form except the
superscript, the way we divide $\mb\beta^{(k)}$ into groups is the same for all
$k$. In the following discussion, $\mb \beta^{(k)}$ will therefore be written as
$\mb \beta$, omitting $k$.

As shown by~(\ref{eqn-seq-linear}), the function $l\,(\x_{1:O},\mb\beta)$ is the
sum of the $\beta$'s associated with the interaction patterns expressed by
$\x_{1:O}$. If a group of interaction patterns are expressed by the same
training cases, the associated $\beta$'s will appear \textit{simultaneously} in
the same factors of~(\ref{eqn-seq-like}). The likelihood
function~(\ref{eqn-seq-like}) therefore depends only on the sum of these
$\beta$'s, rather than the individual ones. Our task is therefore to find the
groups of interaction patterns expressed by the same training cases. 

\begin{figure}[p]

\begin{center}

\includegraphics[scale=0.72]{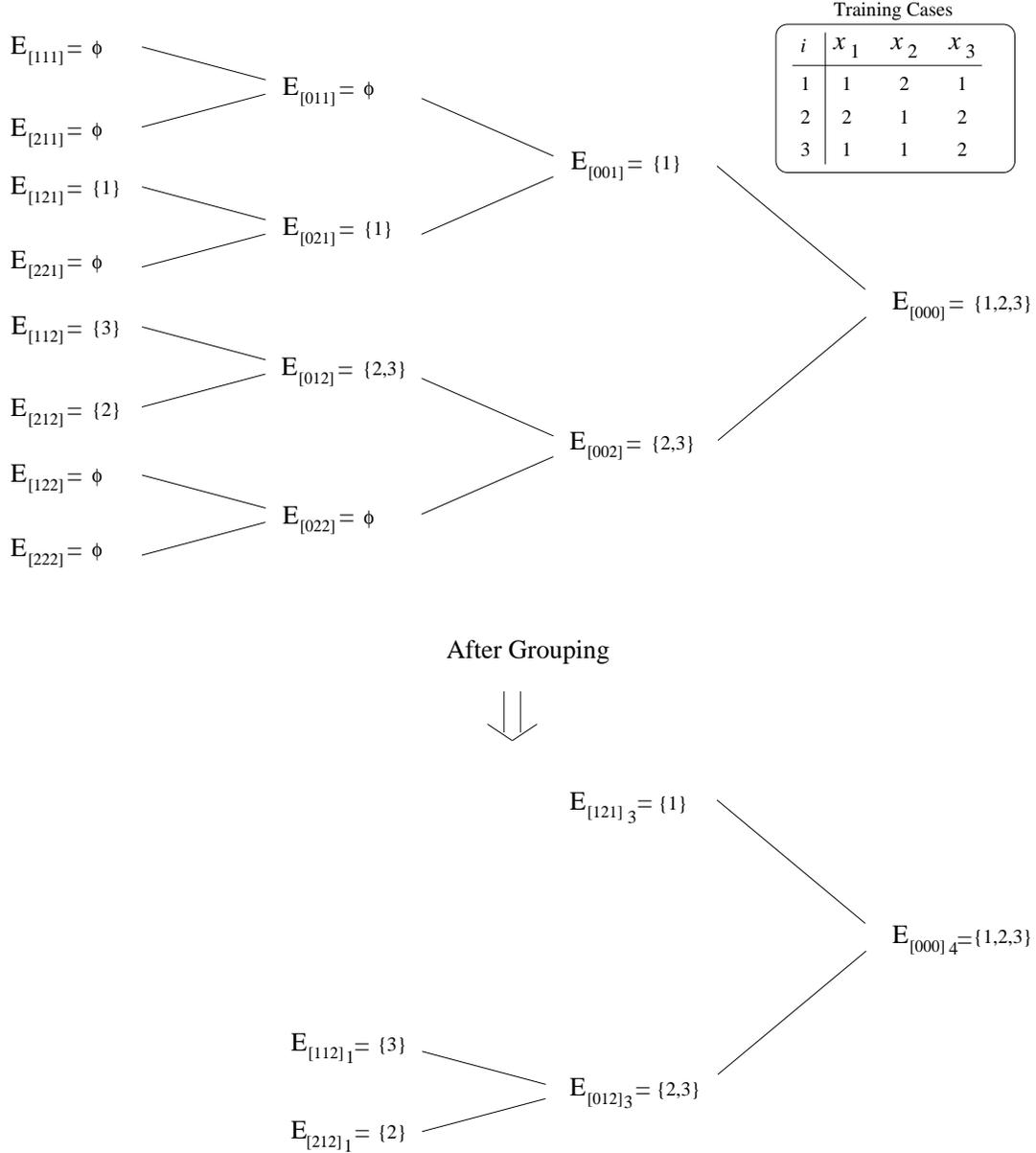}

\end{center}

\caption[A picture showing that the interaction patterns in logistic sequence
prediction models can be grouped, illustrated with binary sequences of length
$O=3$, based on $3$ training cases]{A picture showing that the interaction
patterns in logistic sequence prediction models can be grouped, illustrated with
binary sequences of length $O=3$, based on $3$ training cases shown in the
upper-right box. $E_\P$ is the expression of the pattern (or superpattern) $\P$
--- the indices of the training cases in which the $\P$ is expressed, with
$\phi$ meaning the empty set. We group the patterns with the same expression
together, re-represented collectively by a ``superpattern'', written as
$[0\cdots 0A_b\cdots A_O]_f$,  meaning $\bigcup_{t=b}^{f}\,[0\cdots 0A_t\cdots
A_O]$, where $1 \leq b \leq f \leq O+1$, and in particular when $t=O+1$,
$[0\cdots 0A_t\cdots A_O] = [0\cdots0]$. We also remove the patterns not
expressed by any of the training cases. Only $5$ superpatterns with unique
expressions are left in the lower picture.}

\label{fig-seq-group}

\end{figure}

Let us use $E_{\P}$ to denote the ``expression'' of the pattern $\P$ --- the
indices of training cases in which $\P$ is expressed, a subset of $1,\ldots,N$.
For example, $E_{[0\cdots 0]}=\{1,\ldots,N\}$. In other words, the indicator for
pattern $\P$ has value $1$ for the training cases in $E_{\P}$, and $0$ for
others. We can display $E_\P$ in a tree-shape, as we displayed $\beta_\P$. The
upper part of Figure~\ref{fig-seq-group} shows such expressions for each pattern
of binary sequence of length $O=3$, based on $3$ training cases:
$\x_{1:3}^{(1)}=(1,2,1)$,$\x_{1:3}^{(2)}=(2,1,2)$ and $\x_{1:3}^{(3)}=(1,1,2)$. 
From Figure~\ref{fig-seq-group}, we can see that the expression of a ``stem''
pattern is equal to the union of the expressions of its ``leaf'' patterns, for
example, $E_{[000]}=E_{[001]}\bigcup E_{[002]}$ . 

\begin{figure}[p]

\begin{center}

\includegraphics[scale=0.9]{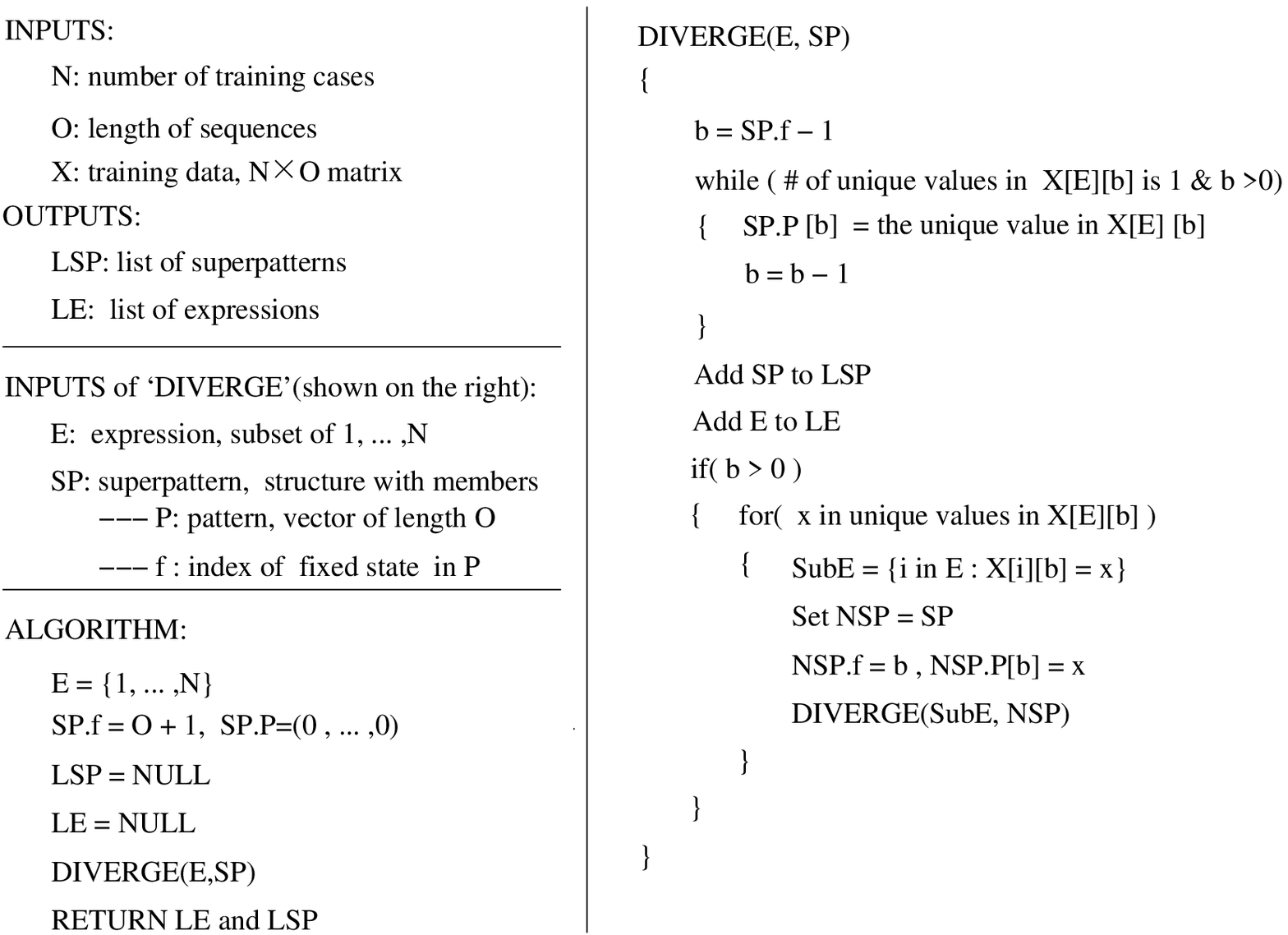}

\end{center}

\caption[The algorithm for grouping parameters of Bayesian logistic sequence
prediction models.]{The algorithm for grouping parameters of Bayesian logistic
sequence prediction models. To group parameters, we call function ``DIVERGE''
with the initial values of expression $E=\{1,\ldots,N\}$ and superpattern
$SP=[0\ldots 0]_{O+1}$, as shown in above picture, resulting in two lists of the
same length, LE and LSP, respectively storing the expressions and the
corresponding superpatterns. Note that the first index of an array is assumed to
be $1$, and that the $X[E][b]$ means a 1-dimension subarray  of $X$ in which the
row indices are in $E$ and the column index equals $b$. }

\label{fig-alg-seq-group}

\end{figure}

When a stem pattern has only one leaf pattern with non-empty expression, the
stem and leaf patterns have the same expression, and can therefore be grouped
together. This grouping procedure will continue by taking the leaf pattern as
the new stem pattern, until encountering a stem pattern that ``splits'', i.e.
has more than one leaf pattern with non-empty expression.  For example,
$E_{[001]},E_{[021]}$ and $E_{[121]}$ in Figure~\ref{fig-seq-group} can be
grouped together. All such patterns must be linked by lines, and can be
represented collectively with a ``superpattern'' $SP$, written as $[0\cdots
0A_b\cdots A_O]_f=\bigcup_{t=f}^{b}\,[0\cdots 0A_t\cdots A_O]$, where $1\leq b
\leq f \leq O+1$, and in particular when $t=O+1$, $[0\cdots 0A_t\cdots A_O] =
[0\cdots0]$. One can easily translate the above discussion into a  computer
algorithm. Figure~\ref{fig-alg-seq-group} describes the algorithm for grouping
parameters of Bayesian logistic sequence prediction models, in a C-like
language, using a recursive function.

\label{sec-comp-blsm}

%%%%%%%%%%%%%%%%%%%%%%%%%%%%%%%%%%%%%%%%%%%%%%%%%%%%%%%%%%%%%%%%%%%%%%%

An important property of our method for compressing parameters of sequence
prediction models is that given $N$ sequences as training data, conceivably of
infinite length, denoted by $x^{(i)}_{-\infty},\ldots,x^{(i)}_{-1}$, for
$i=1,\ldots,N$, the number of superpatterns with unique expressions, and
accordingly the number of compressed parameters, will converge to a finite
number as $O$ increases. The justification of this claim is that if we keep
splitting the expressions following the tree shown in
Figure~\ref{fig-seq-group}, at a certain time, say $t$, every expression will be
an expression with only 1 element (suppose we in advance remove the sequences
that are identical with another one). When considering further smaller $t$, no
more new superpattern with different expressions will be introduced, and the
number of superpatterns will not grow. The number of the \textit{compressed
parameters}, the regression coefficients for the superpatterns, will therefore
not grow after the time $t$.

In contrast, after the time $t$ when each interaction pattern is expressed by
only $1$ training case, if the order is increased by $1$, the number of
interaction patterns is increased by the number of training cases. The
regression coefficients associated with these original interaction patterns,
called \textit{the original parameters} thereafter, will grow linearly with the
order considered. Note that these original parameters do not include the
regression coefficients for those interaction patterns not expressed by any
training case. The total number of regression coefficients defined by  the model
grows exponentially with the order considered.

\subsection{Making Prediction for a Test Case}\label{sec-blsm-pred}

Given $\beta^{(1)},\ldots,\mb \beta^{(K)}$, the predictive probability for the
next state $\x^*_{O+1}$ of a test case for which we know the historic sequence
$\x^*_{1:O}$ can be computed using equation~(\ref{eqn-softmax}), applied to 
$\x^*_{1:O}$. A Monte Carlo estimate of $P(x^*_{O+1} = k\given
\x^*_{1:O},\trainingdata)$ can be obtained by averaging~(\ref{eqn-softmax}) over
the Markov chain samples from the posterior distribution of
$\mb\beta^{(1)},\ldots,\mb\beta^{(K)}$. 

Each of the $O+1$ patterns expressed by the  test case $\x^*_{1:O}$ is either
expressed by some training case (and therefore belongs to one of the
superpatterns), or is a new pattern (not expressed by any training case).
Suppose we have found $\gamma$ superpatterns. The $O+1$ $\beta$'s in the linear
function $l(\x_{1:O}^*,\beta^{(k)})$ can accordingly be divided into $\gamma+1$
groups (some groups may be empty). The function $l(\x_{1:O}^*,\beta^{(k)})$ can
be written as the sum of the sums of the $\beta$'s over these $\gamma+1$ groups.
Consequently, $P(x^*_{O+1} = k\given \x^*_{1:O})$ can be written in the form
of~(\ref{eqn-blsm-pred}). As discussed in Section~\ref{sec-split-pred}, we need
to only split the sum of the $\beta$'s associated with a superpattern, i.e., a
compressed parameter $s_g$, into two parts, such that one of them is the sum of
those $\beta$ expressed by the  test case $\x^*_{1:O}$, using the splitting 
distribution~(\ref{eqn-split2}). 

It is easy to identify the patterns that are also expressed by $\x^*_{1:O}$ from
a superpattern $[0\cdots A_b\cdots A_O]_f$. If
$(x^*_f,\ldots,x^*_O)\not=(A_f,\ldots,A_O)$, none of the patterns in $[0\cdots
A_b\cdots A_O]_f$ are expressed by $\x^*_{1:O}$, otherwise, if
$(x^*_{b'},\ldots,x^*_{O})=(A_{b'},\ldots,A_O)$ for some $b'$ ($b\leq b' \leq
f$), all patterns in $[0\cdots A_{b'}\cdots A_O]_f$ are expressed by
$\x^*_{1:O}$. 

\subsection{Experiments with a Hidden Markov Model} \label{sec-sim-blsm}

In this section we apply Bayesian logistic sequence prediction modeling, with or
without our compression method, to data sets generated using a Hidden Markov
model, to demonstrate our method for compressing parameters. The experiments
show that when the considered length of the sequence $O$ is increased, the
number of compressed parameters will converge to a fixed number, whereas the
number of original parameters will increase linearly. Our compression method
also improves the quality of Markov chain sampling in terms of autocorrelation.
We therefore obtain good predictive performances in a small amount of time using
long historic sequences.

\subsubsection{The Hidden Markov Model Used to Generate the Data}

\begin{figure}[ht]

\begin{center}

\includegraphics[scale=1]{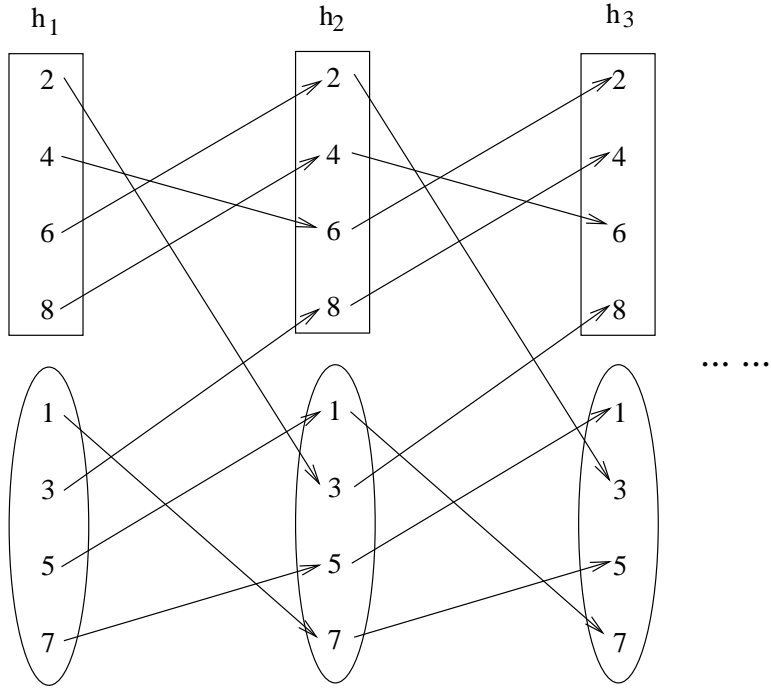}

\end{center}

\caption[A picture showing a Hidden Markov Model, which is used to generate
sequences to demonstrate  Bayesian logistic sequence prediction models]{ A
picture showing a Hidden Markov Model, which is used to generate sequences to
demonstrate  Bayesian logistic sequence prediction models. Only the dominating
transition probabilities of 0.95 are shown using arrows in the above graph,
while from any state the hidden Markov chain can also move to any other state
with a small probability. When $h_t$ is in a rectangle, $x_t$ is equal to $1$
with probability 0.95, and $2$ with probability 0.05, otherwise, when $h_t$ is
in an oval, $x_t$ is equal to $2$ with probability 0.95, and $1$ with
probability 0.05. }

\label{fig-hmm}

\end{figure}

Hidden Markov models (HMM) are applied widely in many areas, for example, speech
recognition (Baker 1975), image analysis (Romberg et.al. 2001), computational
biology (Sun 2006). In a simple hidden Markov model, the observable sequence
$\{x_t\given t=1,2,\ldots\}$ is modeled as a noisy representation  of a hidden
sequence $\{h_t \given t=1,2,\ldots\}$ that has the Markov property (the
distribution of $h_t$ given $h_{t-1}$ is independent with the previous states
before $h_{t-1}$). Figure~\ref{fig-hmm} displays the hidden Markov model used to
generate our data sets, showing the transitions of three successive states. The
hidden sequence $h_t$ is an Markov chain with state space $\{1,\ldots,8\}$,
whose dominating transition probabilities are shown by the arrows in
Figure~\ref{fig-hmm}, each of which is 0.95. However, the hidden Markov chain
can move from any state to any other state as well, with some small 
probabilities. If $h_t$ is an even number, $x_t$ will be equal to $1$ with
probability 0.95 and $2$ with probability 0.05, otherwise, $x_t$ will be equal
to $2$ with probability 0.95 and $1$ with probability 0.05. The sequence
$\{x_t\given t=1,2,\ldots\}$ generated by this exhibits high-order dependency,
though the hidden sequence is only a Markov chain. We can see this by looking at
the transitions of observable $x_t$ in Figure~\ref{fig-hmm}. For example, if
$x_1=1$ (rectangle) and $x_2=2$ (oval), it is most likely to be generated by
$h_1=2$ and $h_2=3$, since this is the only strong connection from the rectangle
to the oval, consequently, $h_3=8$ is most likely to to be the next, and $x_3$
is therefore most likely to be $1$ (rectangle). 

\subsubsection{Experiment Results}

We used the HMM in Figure~\ref{fig-hmm} to generate $5500$ sequences with length
$21$. We used $5000$ sequences as test cases,  and the remaining $500$ as the 
training cases. We tested the prediction methods by predicting $x_{21}$ based on
varying numbers of preceding states, $O$, chosen from the set
$\{1,2,3,4,5,7,12,15,17,20\}$. 

\begin{figure}[p]

\begin{center}

\includegraphics[scale=0.8]{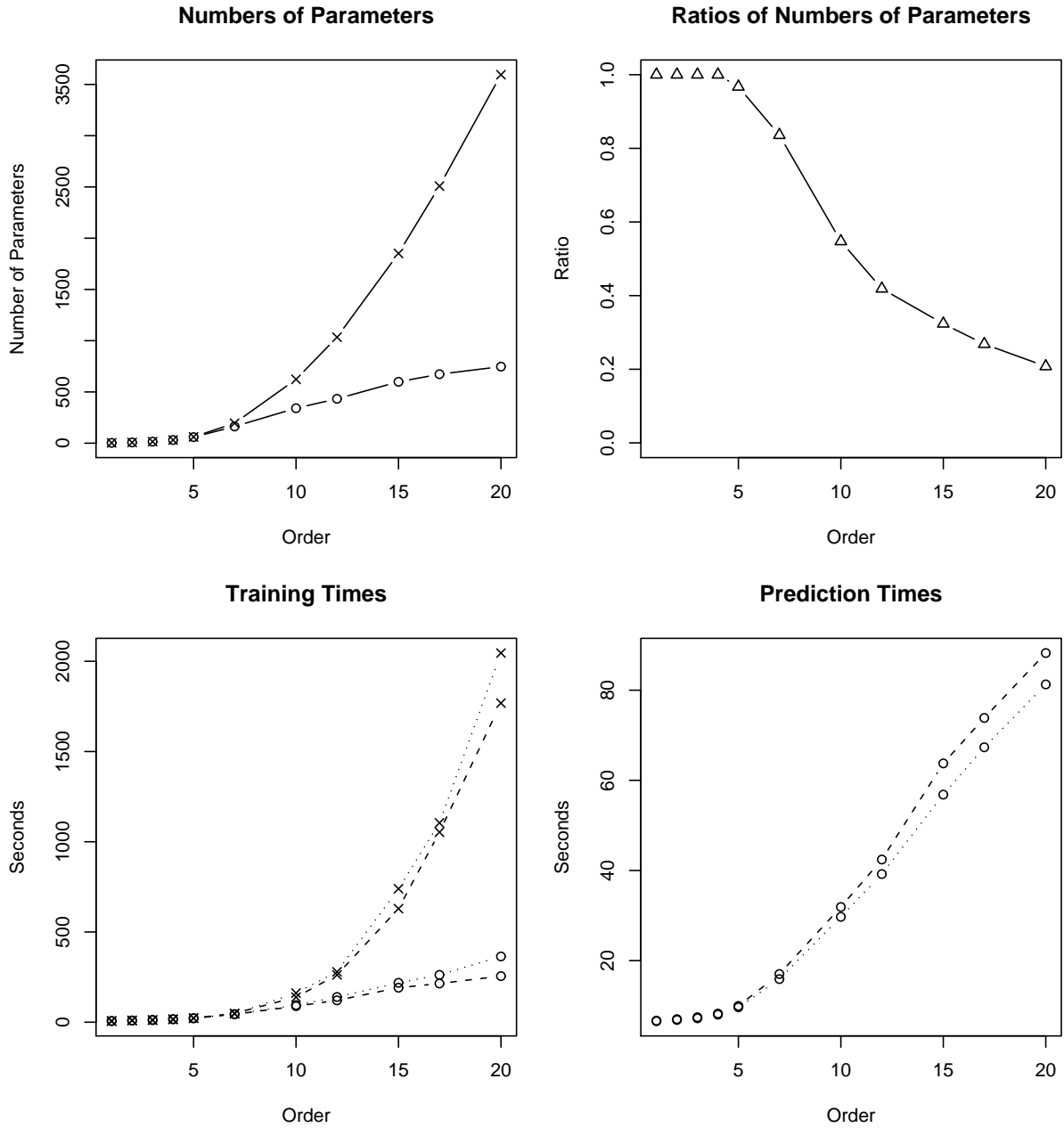}

\end{center}

\caption[Plots showing the reductions of the number of
parameters and the training time with our compression method using the
experiments on a data set generated by a HMM]{Plots showing the reductions of
the number of parameters and the training time with our compression method using
the experiments on a data set generated by a HMM. The upper-left plot shows the
number of the compressed and the original parameters based on $500$ training
sequences for $O=1,2,3,4,5,7,10,12,15,17,20$, their ratios are shown in the
upper-right plot. In the above plots, the lines with $\circ$ are for the methods
with parameters compressed, the lines with $\times$ are for the methods without
parameters compressed, the dashed lines are for the methods with Gaussian
priors, and the dotted lines are for the methods with Cauchy priors. The
lower-left plot shows the training times for the methods with and without
parameters compressed. The lower-right plot shows the prediction time only for 
the methods with parameters compressed. }

\label{fig-hmm500-comp}

\end{figure}

Figure~\ref{fig-hmm500-comp} compares the number of parameters and the times
used to train the model, with and without our compression method.  It is clear
that our method for compressing parameters reduces greatly the number of
parameters. The ratio of the number of compressed parameters to the number of
the original ones decreases with the number of preceding states, $O$. For
example, the ratio reaches $0.207$ when $O=20$. This ratio will reduce to $0$
when considering even bigger $O$, since the number of original parameters will
grow with $O$ while the number of compressed parameters will converge to a
finite number, as discussed in Section~\ref{sec-comp-blsm}. There are similar
reductions for the training times with our compression method. But the training
time with compressed parameters will not converge to a finite amount, since the
time used to update the hyperparameters ($\sigma_o$'s) grows with order, $O$.
Figure~\ref{fig-hmm500-comp} also shows the prediction times for $5000$ training
cases. The small prediction times show that the methods for splitting Gaussian
and Cauchy variables are very fast. The prediction times grow with $O$ because
the time used to identify the patterns in a superpattern expressed by a test
case grows with $O$. The prediction times with the original parameters are not
shown in Figure~\ref{fig-hmm500-comp}, since we do not claim that our
compression method saves prediction time. (If we used the time-optimal 
programming method for each method, the prediction times with compressed
parameters should be more than without compressing parameters since the method
with compression should include times for identifying the patterns from the
superpattern for test cases. With our software, however, prediction times with
compression are less than without compression, which is not shown in
Figure~\ref{fig-hmm500-comp}, because the method without compression needs to
repeatedly read a huge number of the original parameters into memory from disk.)

%%%%%%%%%%%%%%%%%%%%%%%%%%%%%%%%%%%%%%%%%%%%%%%%%%%%%%%%%%%%%%%%%%%%%%%

%%%%%%%%%%%%%%%%%%%%%%%%%%%%%%%%%%%%%%%%%%%%%%%%%%%%%%%%%%%%%%%%%%%%%%%
\begin{figure}[p]

\begin{center}

\includegraphics[scale=0.65]{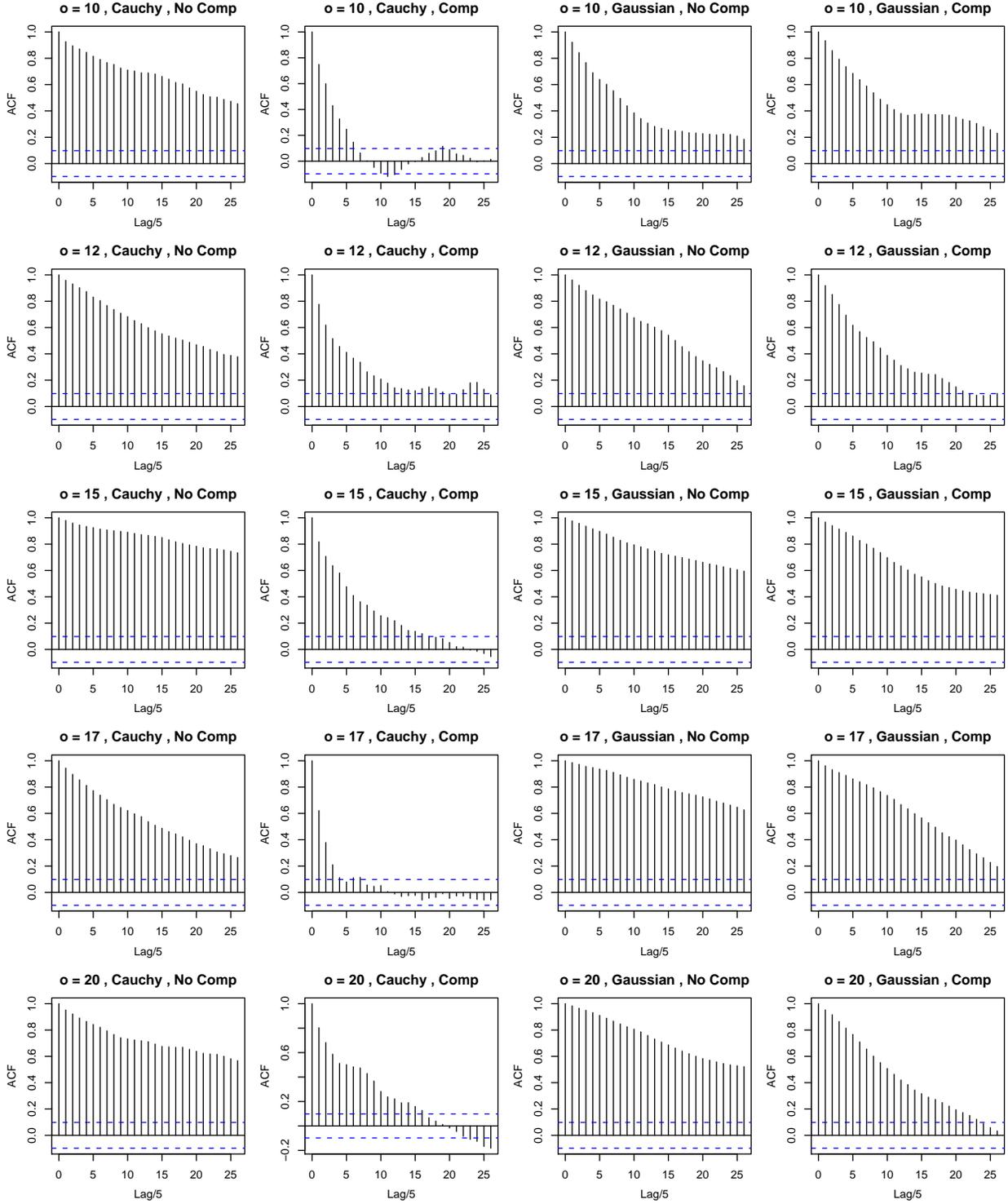}

\end{center}

\caption[The autocorrelation plots of the Markov chains of
$\sigma_o$'s for the experiments on a data set generated by a HMM]{The
autocorrelation plots of $\sigma_o$'s for the experiments
on a data set generated by a HMM, when the length of the preceding sequence
$O=20$. We show the autocorrelations of $\sigma_o$, for $o=10,12,15,17,20$. In
the above plots, ``Gaussian'' in the titles indicates the methods with Gaussian
priors, ``Cauchy'' indicates with Cauchy priors, ``comp'' indicates with
parameters compressed, ``no comp'' indicates without parameters compressed.}

\label{fig-hmm500-acf}

\end{figure}

%%%%%%%%%%%%%%%%%%%%%%%%%%%%%%%%%%%%%%%%%%%%%%%%%%%%%%%%%%%%%%%%%%%%%%%
\begin{figure}[p]

\begin{center}

\includegraphics[height=6in,width=6.4in]{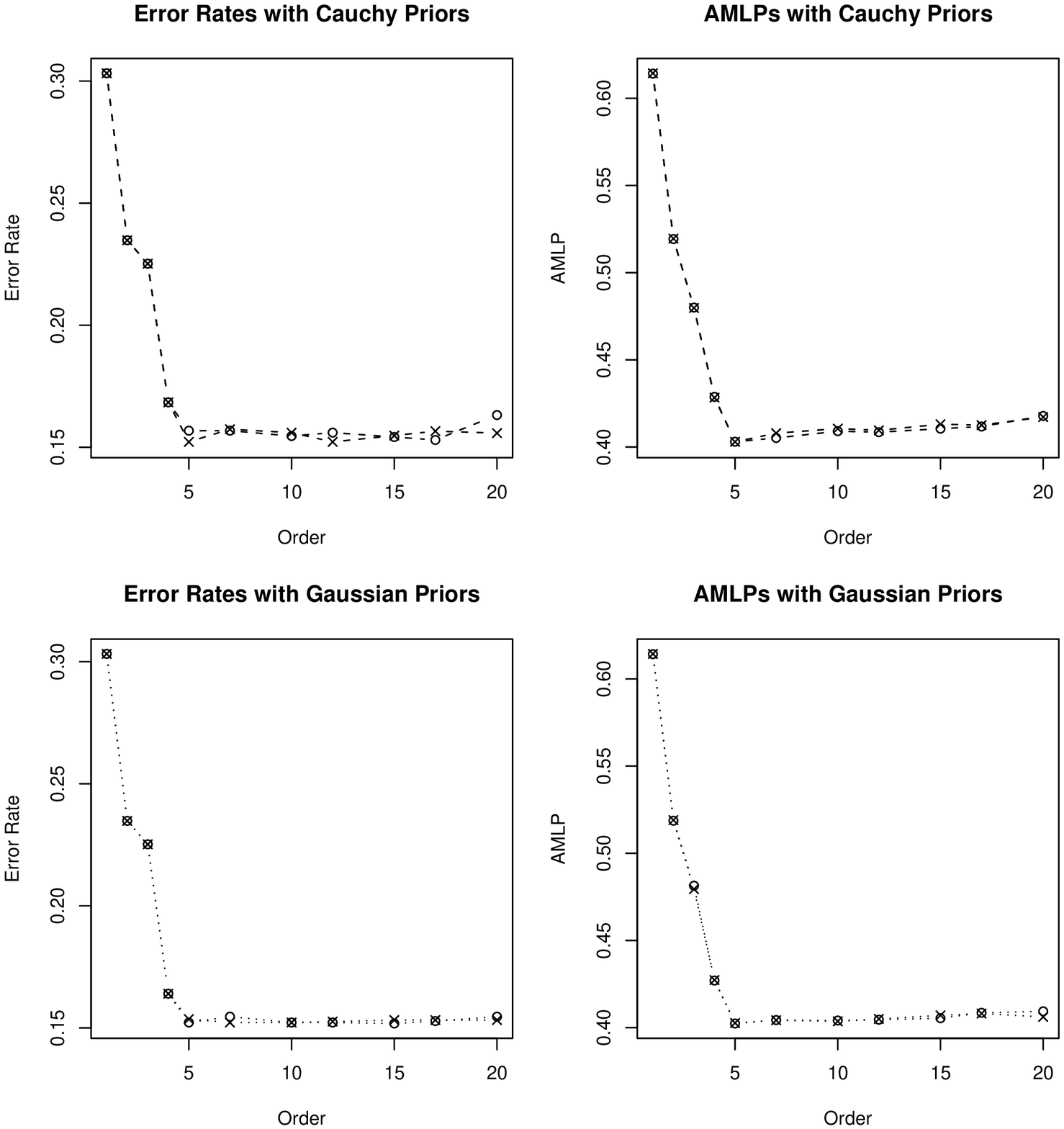}

\end{center}

\caption[Plots showing the predictive performance using the experiments on a
data set generated by a HMM]{Plots showing the predictive performance using the
experiments on a data set generated by a HMM. The left plots show the error
rates and the right plots show the average minus log probabilities of the true
responses in the test cases. The upper plots show the results when using the
Cauchy priors and the lower plots shows the results when using the Gaussian
priors. In all plots, the lines with $\circ$ are for the methods with parameters
compressed, the lines with $\times$ are for the methods without parameters
compressed. The numbers of the training and test cases are respectively $500$
and $5000$. The number of classes of the response is $2$.}

\label{fig-hmm500-error}

\end{figure}

Compressing parameters also improves the quality of Markov chain sampling.
Figure~\ref{fig-hmm500-acf} shows the autocorrelation plots of the
hyperparameters $\sigma_o$, for $o=10,12,15,17,20$, when the length of the
preceding sequence, $O$, is $20$. It is clear that the autocorrelation decreases
more rapidly with lag when we compress the parameters. This results from the
compressed parameters capturing the important directions of the likelihood
function (i.e. the directions where a small change can result in large a change
of the likelihood). We did not take the time reduction from compressing
parameters into consideration in this comparison. If we rescaled the lags in the
autocorrelation plots according to the computation time, the reduction of
autocorrelation of Markov chains with the compressed parameters would be much
more pronounced. 

Finally, we evaluated the predictive performance in terms of error rate (the
fraction of wrong predictions in test cases), and the average minus log
probability (AMLP) of observing the true response in a test case based on the
predictive probability for different classes. The performance of with and
without compressing parameters are the same, as should be the case in theory,
and will be in practice when the Markov chains for the two methods converge to
the same modes. Performance of methods with Cauchy and Gaussian priors is also
similar for this example. The predictive performance is improved when $O$ goes
from $1$ to $5$. When $O > 5$ the predictions are slightly worse than with $O=5$
in terms of AMLP. The error rates for $O>5$ are almost the same as for $O=5$.
This shows that the Bayesian models can perform reasonably well even when we
consider a very high order, as they avoid the overfitting problem in using
complex models. We therefore do not need to restrict the order of the Bayesian
sequence prediction models to a very small number, especially after applying our
method for compressing parameters.

%%%%%%%%%%%%%%%%%%%%%%%%%%%%%%%%%%%%%%%%%%%%%%%%%%%%%%%%%%%%%%%%%%%%%%%

\subsection{Experiments with English Text} \label{sec-text-blsm}

We also tested our method using a data set created from an online article from
the website of the Department of Statitics, University of Toronto. In creating
the data set, we encoded each character as $1$ for vowel letters (a,e,i,o,u),
$2$ for consonant letters, and $3$ for all other characters, such as space,
numbers, special symbols, and we then collapsed multiple occurrences of ``$3$''
into only $1$ occurrence. The length of the whole sequence is 3930. Using it we
created a data set with $3910$ overlaped sequences of length $21$, and used the
first $1000$ as training data. 

%%%%%%%%%%%%%%%%%%%%%%%%%%%%%%%%%%%%%%%%%%%%%%%%%%%%%%%%%%%%%%%%%%%%%%%
\begin{figure}[p]

\begin{center}

\includegraphics[scale=0.8]{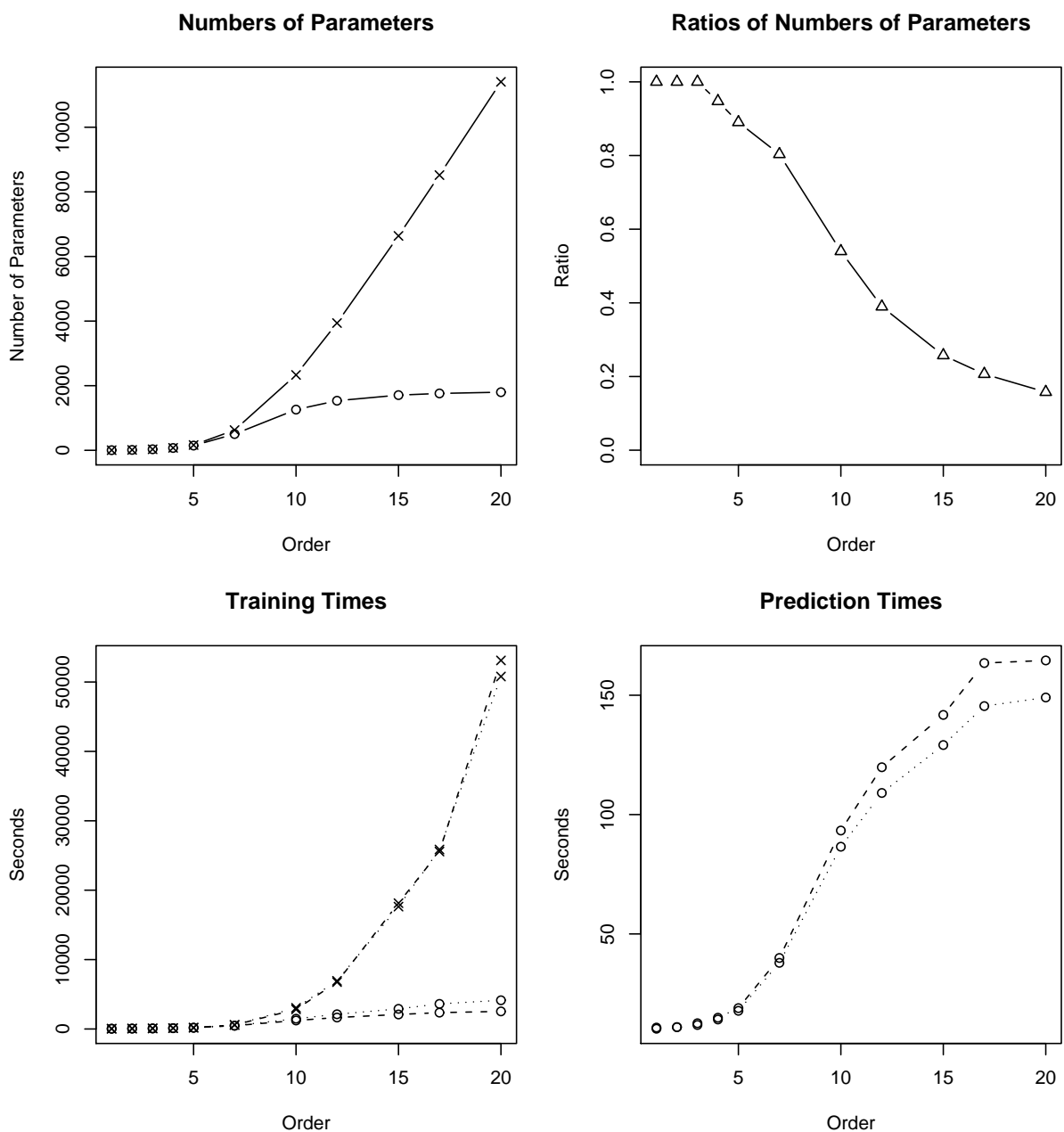}

\end{center}

\caption[Plots showing the reductions of the number of
parameters and the training time with our compression method using the
experiments on English text]{Plots showing the reductions of the number of
parameters and the training and prediction time with our compression method
using the experiments on English text. The upper-left plot shows the number of
the compressed and the original parameters based on $500$ training sequences for
$O=1,2,3,4,5,7,10,12,15,17,20$, their ratios are shown in the upper-right plot.
In the above plots, the lines with $\circ$ are for the methods with parameters
compressed, the lines with $\times$ are for the methods without parameters
compressed, the dashed lines are for the methods with Gaussian priors, and the
dotted lines are for the methods with Cauchy priors. The lower-left plot shows
the training times for the methods with and without parameters compressed. The
lower-right plot shows the prediction time only for the methods with parameters
compressed. }

\label{fig-text-comp}

\end{figure}

%%%%%%%%%%%%%%%%%%%%%%%%%%%%%%%%%%%%%%%%%%%%%%%%%%%%%%%%%%%%%%%%%%%%%%%
\begin{figure}[p]

\begin{center}

\includegraphics[scale=0.65]{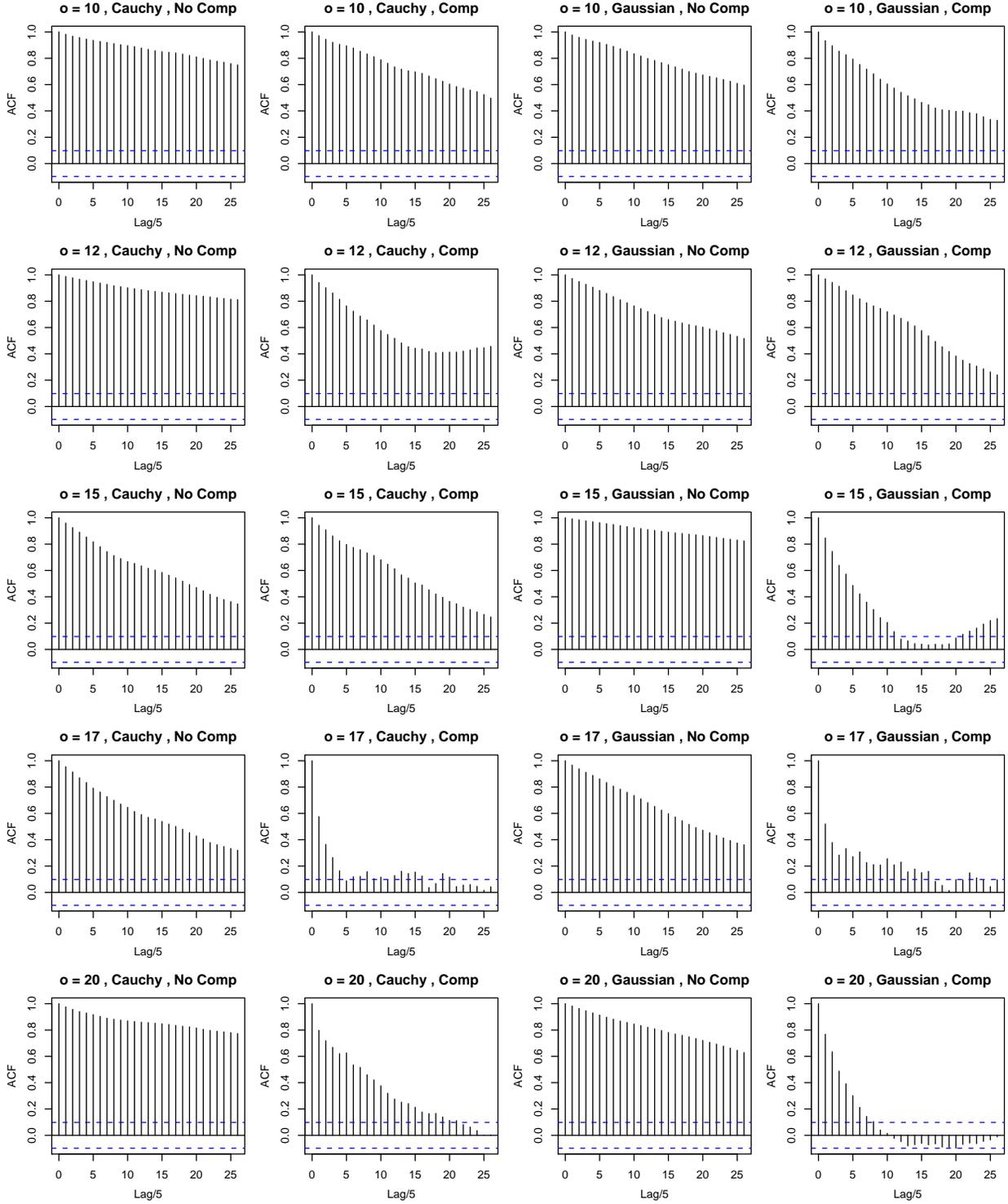}

\end{center}

\caption[The autocorrelation plots of the $\sigma_o$'s for the
experiments on English text data]{The autocorrelation plots of the $\sigma_o$'s
for the experiments on English text data, when the length of the preceding
sequence $O=20$. We show the autocorrelation plot of $\sigma_o$, for
$o=10,12,15,17,20$. In the above plots, ``Gaussian'' in the titles indicates the
methods with Gaussian priors, ``Cauchy'' indicates with Cauchy priors, ``comp''
indicates with parameters compressed, ``no comp'' indicates without parameters
compressed.}

\label{fig-text-acf}

\end{figure}

%%%%%%%%%%%%%%%%%%%%%%%%%%%%%%%%%%%%%%%%%%%%%%%%%%%%%%%%%%%%%%%%%%%%%%%
\begin{figure}[p]

\begin{center}

\includegraphics[height=6in,width=6.4in]{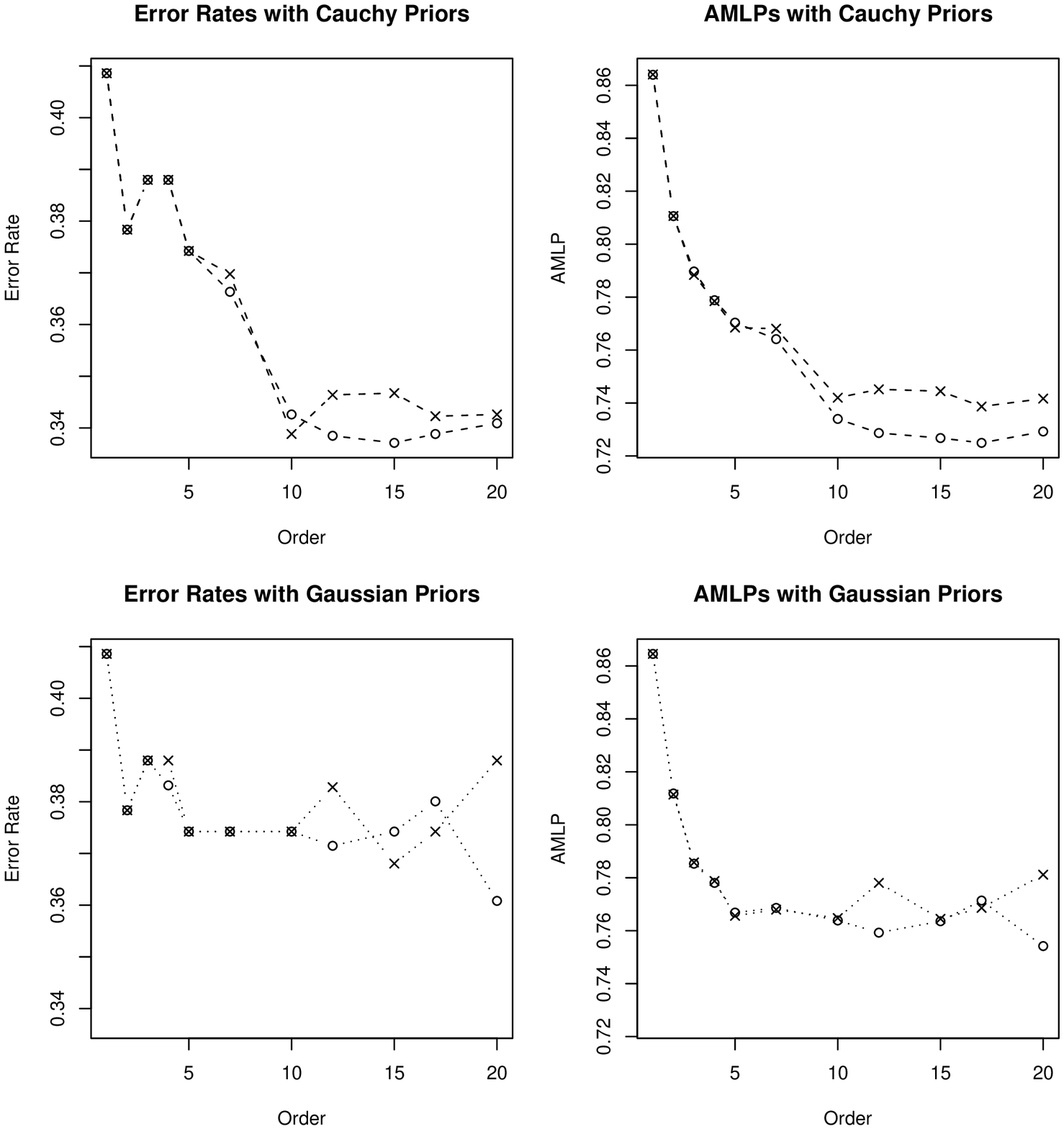}

\end{center}

\caption[Plots showing the predictive performance using the experiments on
English text data]{Plots showing the predictive performance using the
experiments on English text data. The left plots show the error rate and the
right plots show the average minus log probability of the true response in
a test case. The upper plots show the results when using the Cauchy priors
and the lower plots shows the results when using the Gaussian priors. In all
plots, the lines with $\circ$ are for the methods with parameters compressed,
the lines with $\times$ are for the methods without parameters compressed. The
numbers of the training and test cases are respectively $1000$ and $2910$. The
number of classes of the response is $3$.}

\label{fig-text-error}

\end{figure}

\begin{figure}[t]

\begin{center}

\includegraphics[scale=0.65]{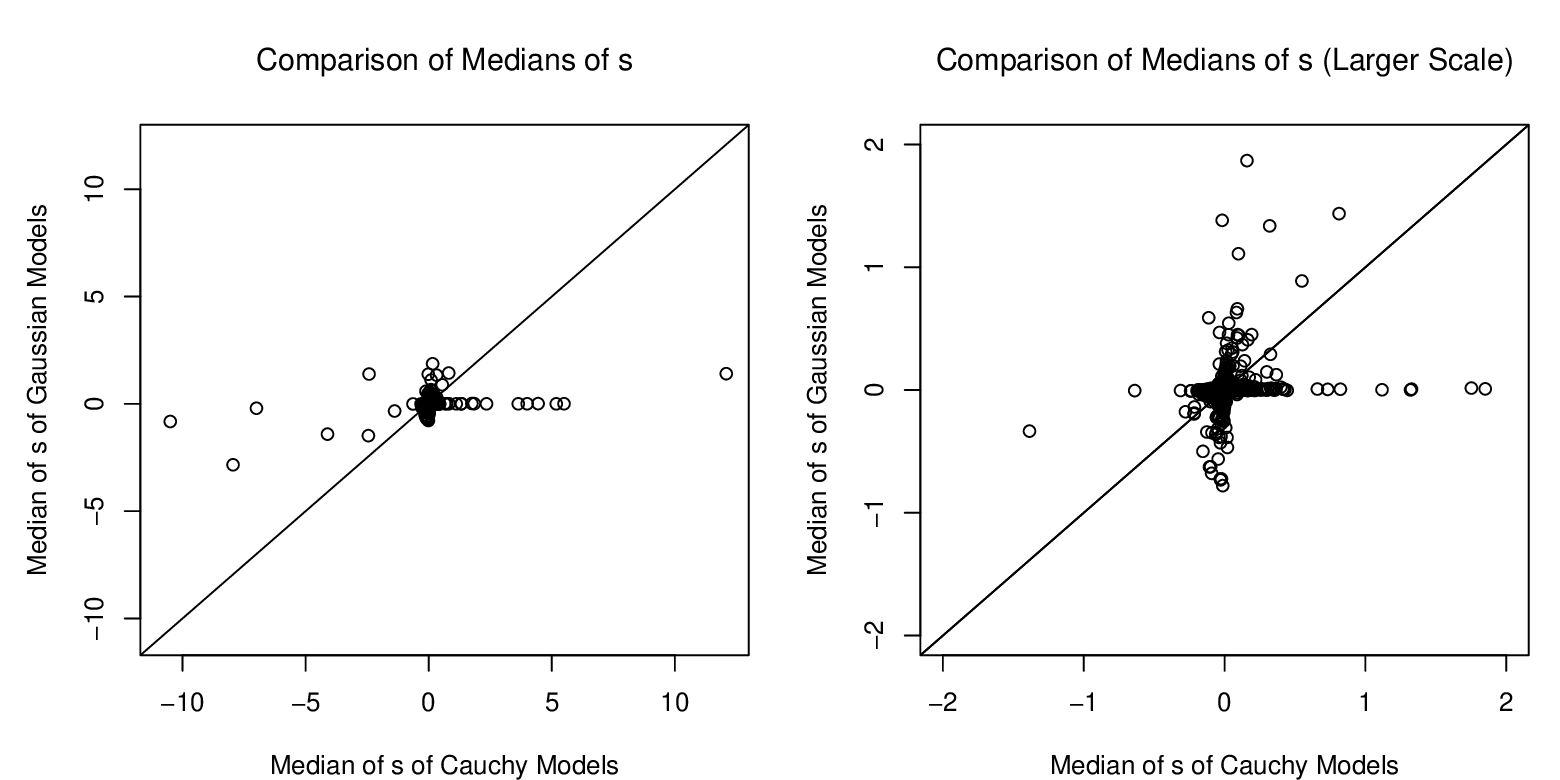}

\end{center}

\caption[Scatterplots of the medians of all the compressed parameters,for the
models with Cauchy and Gaussian priors, fitted with English text
data]{Scatterplots of medians of all compressed parameters, $s$, of Markov chain
samples in the last $1250$ iterations, for the models with Cauchy and Gaussian
priors, fitted with English text data, with the length of preceding sequence
$O=10$, and with the parameters compressed. The right plot shows in a larger
scale the rectangle $(-2,2)\times (-2,2)$. }

\label{fig-text-medians}

\end{figure}

The experiments were similar to those in Section~\ref{sec-sim-blsm}, with the
same priors and the same computational specifications for Markov chain sampling.
Figures~\ref{fig-text-comp},  \ref{fig-text-acf}, \ref{fig-text-error}, and
\ref{fig-text-medians}  show the results. All the conclusions drawn from the
experiments in Section~\ref{sec-sim-blsm} are confirmed in this example, with
some differences in details. In summary, our compression method reduces greatly
the number of parameters, and therefore shortens the training process greatly.
The quality of Markov chain sampling is improved by compressing parameters.
Prediction is very fast using our splitting methods. The predictions on the test
cases are improved by considering higher order interactions. From
Figure~\ref{fig-text-error}, at least some order $10$ interactions are useful in
predicting the next character.

In this example we also see that when Cauchy priors are used Markov chain
sampling with the original parameters may have been trapped in a local mode,
resulting in slightly worse predictions on test cases than with the compressed
parameters, even though the models used are identical.

%%%%%%%%%%%%%%%%%%%%%%%%%%%%%%%%%%%%%%%%%%%%%%%%%%%%%%%%%%%%%%%%%%%%%%%

%%%%%%%%%%%%%%%%%%%%%%%%%%%%%%%%%%%%%%%%%%%%%%%%%%%%%%%%%%%%%%%%%%%%%%%

\begin{figure}[p]

\begin{center}

\includegraphics[width=6.5in]{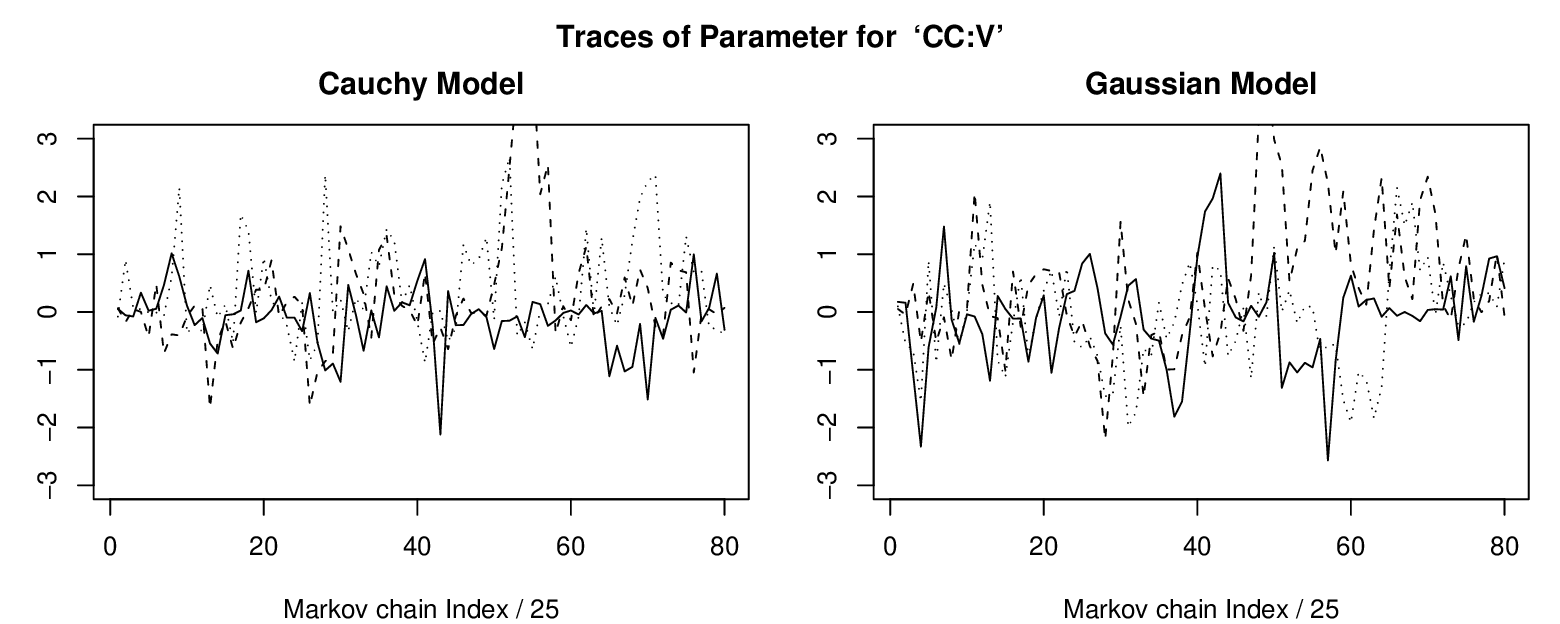}

\vspace*{0.1in}

\hrule

%\vspace*{0.1in}

\includegraphics[width=6.5in]{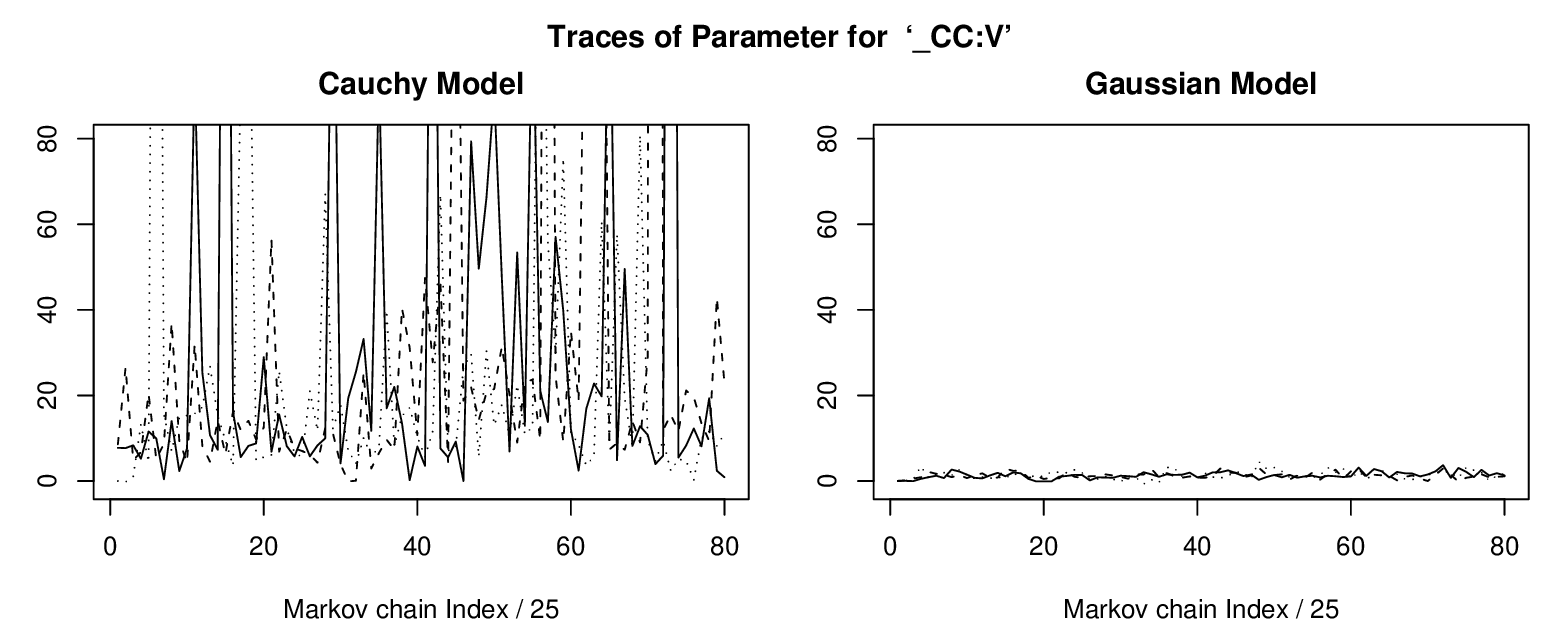}

\vspace*{0.1in}

\hrule

%\vspace*{0.1in}

\includegraphics[width=6.5in]{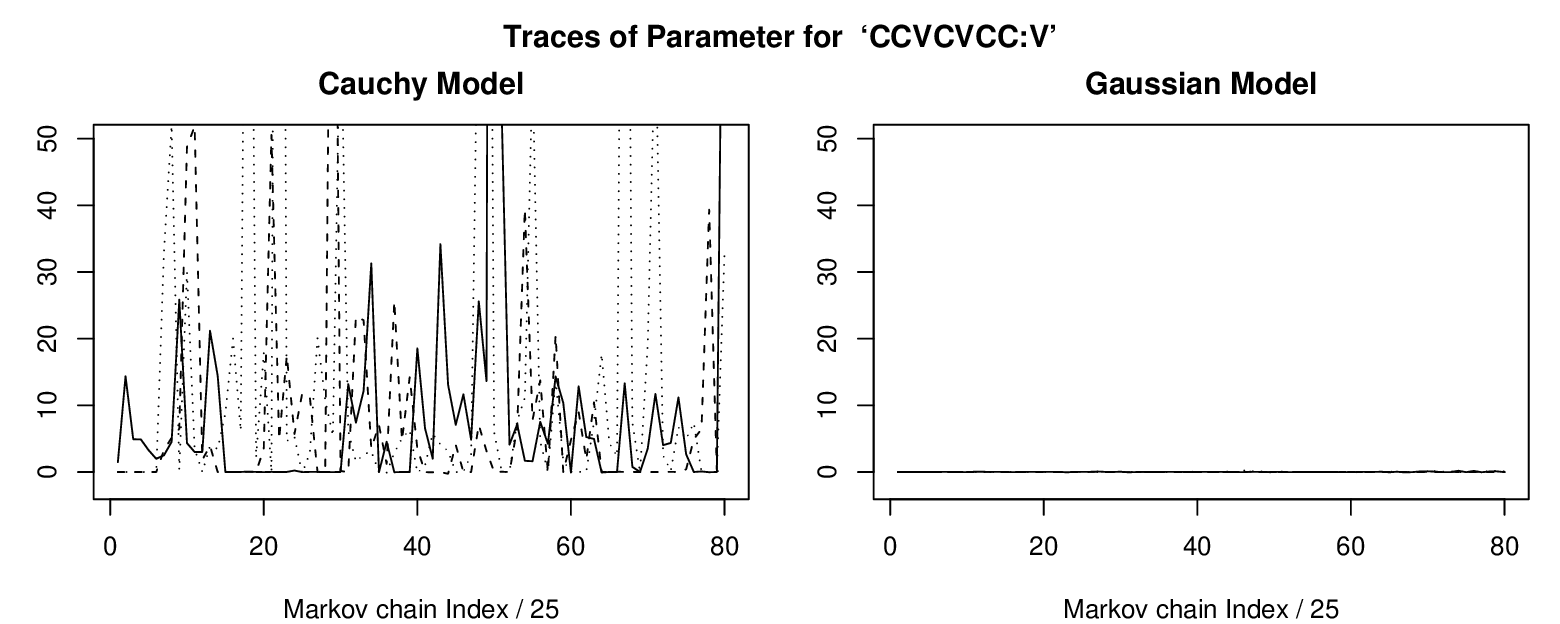}

\end{center}

\caption[Plots of Markov chain traces of three compressed
parameters from Experiments on English text]{Plots of Markov chain traces of
three compressed parameters (Each contains only one $\beta$) from Experiments on
English text with $10$ preceding states, with Cauchy or Gaussian priors. The
parameters are annotated by their original meanings in English sequence. For
example, `\_\_CC:V' stands for the parameter for predicting that the next
character is a ``vowel'' given preceding three characters are ``space,
consonant, consonant''.}

\label{fig-text-betas}

\end{figure}

We also see that the models with Cauchy priors result in better predictions than
those with Gaussian priors for this data set, as seen from the plots of error
rates and AMLPs. To investigate the difference of using Gaussian and Cauchy
priors, we first plotted the medians of Markov chains samples (in the last
$1250$ iteractions) of all compressed parameters, $s$, for the model with
$O=10$, shown in Figure~\ref{fig-text-medians}, where the right plot shows in a
larger scale the rectangle $(-2,2)\times(-2,2)$.  This figure shows that a few
$\beta$ with large medians in the Cauchy model have very small corresponding
medians in the Gaussian model.

We also looked at the traces of some compressed parameters, as shown in
Figure~\ref{fig-text-betas}. The three compressed parameters shown all contain
only a single $\beta$. The plots on the top are for the $\beta$ for ``CC:V'',
used for predicting whether the next character is a vowel given the preceding
two characters are consonants; the plots in the middle are for ``\_\_CC:V",
where ``\_\_'' denotes a space or special symbol; the plots on the bottom are 
for ``CCVCVCC:V'', which had the largest median among all compressed parameters
in the Cauchy model, as shown by Figure~\ref{fig-text-medians}. The regression
coefficient $\beta$ for ``CC:V'' should be close to $0$ by our common sense,
since two consonants can be followed by any of three types of characters. We can
very commonly see ``CCV'', such as ``the'',  and  ``CC\_\_'', such as
``with\_\_'', and not uncommonly see ``CCC'', such as ``technique'',``world'',
etc. The Markov chain trace of this $\beta$ with a Cauchy prior moves in a
smaller region around $0$ than with a Gaussian prior. But if we look back one
more character, things are different. The regression coefficient $\beta$ for
``\_\_CC:V" is fairly large, which is not surprising. The two consonants in
``\_\_CC:V" stand for two letters in the beginning of a word. We rarely see a
word starting with three consonants or a word consisting of only two consonants.
The posterior distribution of this $\theta$ for both Cauchy and Gaussian models
favor positive values, but the Markov chain trace for the Cauchy model can move
to much larger values than for the Gaussian model. As for the high-order pattern
``CCVCVCC'', it matches words like  ``statistics'' or ``statistical'', which
repeatedly appear in an article introducing a statisics department. Again, the
Markov chain trace of this $\beta$ for the Cauchy model can move to much larger
values than for Gaussian model, but sometimes it is close to $0$, indicating
that there might be two modes for its posterior distributution.

The above investigation reveals that a Cauchy model allows some useful $\beta$
to be much larger in absolute value than others while keeping the useless
$\beta$ in a smaller region around $0$ than a Gaussian model. In other words,
Cauchy models are more powerful in finding the information from the many
possible high-order interactions than Gaussian models, due to the heavy
two-sided tails of Cauchy distributions.

\section{Conclusion and Discussion}\label{sec-comp-conclude}

In this paper, we have proposed a method to effectively reduce the number of
parameters of Bayesian classification and regression models with high-order
interactions, using a compressed parameter to represent the sum of all the
regression coefficients for the predictor variables that have the same values
for all the training cases. Working with these compressed parameters, we greatly
shorten the training time with MCMC. These compressed parameters can later be
split into the original parameters efficiently. We have demonstrated,
theoretically and empirically, that given a data set with fixed number of cases,
the number of compressed parameters will have converged before considering the
highest possible order. Applying Bayesian methods to regression and
classification models with high-order interactions therefore become much easier
after compressing the parameters, as shown by our experiments with simulated and
real data. The predictive performance will be improved by considering high-order
interactions if some useful high-order interactions do exist in the data.

We have devised an efficient scheme for compressing parameters of Bayesian logistic sequence prediction models, as described in Section~\ref{sec-blsm}. The algorithm for sequence prediction models is efficient. The resulting groups of interaction patterns have unique expressions. We have also found similar schemes for compressing parameters of general Bayesian classification models with discrete features, though it is more difficult, see (Li 2007).

We have also empirically demonstrated that Cauchy distributions with location
parameter $0$, which have heavy two-sided tails, are more appropriate than
Gaussian distributions in capturing the prior belief that most of the parameters
in a large group are very close to $0$ but a few of them may be much larger in
absolute value, as we may often think appropriate for the regression
coefficients in certain problems.

We have implemented the compression method only for classification models in
which the response and the features are both discrete. Without any difficulty,
the compression method can be used in regression models in which the response is
continuous but the features are discrete, for which we need only use another 
distribution to model the continuous response variable, for example, a Gaussian
distribution. Unless one converts the continuous features into discrete values,
it is not clear how to apply the method described in this paper to continuous
features. However it seems possible to apply the more general idea that we need
to work only with those parameters that matter in the likelihood function when
training models with MCMC, probably by transforming the original parameters.

\section*{Acknowledgements}

This research was supported by Natural Sciences and Engineering Research 
Council of Canada. Radford Neal holds a Canada Research Chair in 
Statistics and Machine Learning.

\section*{References}

\begin{description}

\item[] Baker, J.~K.\ (1975), ``The Dragon system - an overview'', \textit{IEEE
Transactions on. Acoustic Speech Signal Processing} ASSP-23(1): 24-29.

\item []
  Bell, T.~C., Cleary, J.~G., and Witten, I.~H. (1990), \textit{Text Compression},
  Prentice-Hall

\item[]
  Feller, W.\ (1966), ``An Introduction to Probability Theory and its                        Applications'', Volume II, New York: John Wiley

\item[]
Li, L.\ (2007), Bayesian Classification and Regression with High Dimensional Features, Ph.D. Thesis, University of Toronto, available from http://math.usask.ca/$\sim$longhai.

\item[]
  Neal, R.~M. \ (2003), ``Slice Sampling'', \textit{Annals of Statistics}, 
  vol. 31, p.~705-767

\item[]
  Ritchie, M.~D., Hahn, L.~W., Roodi, N., Bailey, L.~R., Dupont,W.~D.,  
  Parl,F.~F.,\ and Moore, J.H.\ (2001),
  ``Multifactor-Dimensionality Reduction Reveals High-Order Interactions
   among Estrogen-Metabolism Genes in Sporadic Breast Cancer'', \textit{The
   American Journal of Human Genetics}, volume 69, pages 138-147 

\item[] 

   Romberg, J., Choi, H. and Baraniuk, R. (2001), ``Bayesian
   tree-structured image modeling using wavelet-domain hidden Markov models'',
   \textit{IEEE Transactions on image processing} 10(7): 1056-1068.

\item[]

  Sun, S. (2006), ``Haplotype Inference Using a Hidden Markov Model
  with Efficient Markov Chain Sampling'', Ph.D. Thesis, University of Toronto

\item[]
 Thisted, R.~A.\ (1988), \textit{Elements of Statistical Computing}, 
  Chapman and Hall.

\end{description}

\end{document}

%% file: newcommands.tex
\newcommand{\given}{\ | \ }

\newcommand{\itb}{\begin{itemize}}
\newcommand{\itn}{\end{itemize}}
\newcommand{\nmn}{\end{enumerate}}
\newcommand{\nmb}{\begin{enumerate}}
\newcommand{\ctb}{\begin{center}}
\newcommand{\ctn}{\end{center}}
\newcommand{\tb}{\begin{center}\Large \bf}
\newcommand{\tn}{\end{center}}
\newcommand{\arb}{\begin{array}}
\newcommand{\qrn}{\end{array}}
\newcommand{\tblb}{\begin{tabular}}
\newcommand{\tbln}{\end{tabular}}
\newcommand{\fb}{\begin{figure}}
\newcommand{\fn}{\end{figure}}

\def\beq{\noindent \begin{eqnarray}}
\def\eeq{\end{eqnarray}}

\def\beqs {\noindent \begin{eqnarray*}}
\def\eeqs {\end{eqnarray*}}

\def\mb#1{\mbox{\boldmath $#1$}}

\def\x{\mb x}

\def \cc {\mbox{Cauchy}}

\def \lldots {\ \ldots \ }

\def \trainingdata {\mathcal{D}}

\def \P {\mathcal{P}}

\def \sigmaa {\Sigma_1}
\def \sigmab {\Sigma_2}

\def \s {s_g}
\def \st {s^t_g}